\documentclass[iicol]{sn-jnl}
\usepackage{lmodern}
\newcommand{\orcidlink}[1]{\,\href{https://orcid.org/#1}{\includegraphics[height=8.5pt]{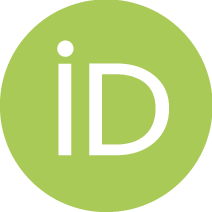}}}
\usepackage{graphicx}%
\usepackage[export]{adjustbox}
\usepackage{multirow}%
\usepackage{amsmath,amssymb,amsfonts}%
\usepackage{amsthm}%
\usepackage{multirow}
\usepackage{enumitem}
\usepackage{balance}
\usepackage{subcaption}
\usepackage{cite}
\usepackage[linesnumbered,ruled,vlined]{algorithm2e}
\usepackage{array}
\usepackage{lipsum}
\usepackage{xcolor}
\usepackage{soul}
\usepackage{booktabs}
\usepackage{multirow}

\newcommand{\PreserveBackslash}[1]{\let\temp=\\#1\let\\=\temp}
\newcolumntype{C}[1]{>{\PreserveBackslash\centering}m{#1}}

\begin{document}

\title[Article Title]{Polar Coordinate-based Differential Evolution for Moving Target Search Using Vision Sensor on Unmanned Aerial Vehicles}

\author[1]{\fnm{Thu Hang} \sur{Khuat}\orcidlink{0009-0003-9611-8678}}

\author[1]{\fnm{Duy-Nam} \sur{Bui}\orcidlink{0009-0001-0837-4360}}

\author[2]{\fnm{Thuy Ngan} \sur{Duong}\orcidlink{0009-0005-9768-7084}}

\author*[3]{\fnm{Manh Duong} \sur{Phung}\orcidlink{0000-0001-5247-6180}}\email{duong.pm@vinuni.edu.vn}

\affil[1]{\orgdiv{VNU University of Engineering and Technology}, \orgname{Vietnam National University, Hanoi}, \orgaddress{\country{Vietnam}}}

\affil[2]{\orgdiv{Department of Electrical Engineering}, \orgname{Ulsan National Institute of Science and Technology}, \orgaddress{\city{Ulsan}, \country{Korea}}}

\affil[3]{\orgdiv{College of Engineering and Computer Science and Smart Green Transformation Center (GREEN-X)}, \orgname{VinUniversity}, \orgaddress{\city{Hanoi}, \country{Vietnam}}}


\abstract{In search and rescue operations, there is a period known as the ``golden time'' during which the probability of finding the target alive is highest. The objective of this work is to propose a new search algorithm for unmanned aerial vehicles (UAVs) with a focus on improving the detection probability and execution time. We approach this problem by first modeling target dynamics as a Markov process and the detection likelihood as a function of image quality and the observer's vision. We then employ Bayesian theory to derive a fitness function representing the probability distribution of the target’s location over the search area. Finally, we introduce a new algorithm named Polar coordinate-based Differential Evolution (PDE) to generate a UAV search path that maximizes this fitness function. The PDE algorithm utilizes polar coordinates to incorporate kinematic constraints and maneuver properties of the UAV, allowing for better exploration of the solution space. A series of simulations and comparative analyses have been conducted to evaluate the performance of the proposed algorithm. Experiments involving a real UAV have also been conducted. Results demonstrate that the PDE algorithm outperforms state-of-the-art algorithms in terms of detection probability and execution time across diverse search scenarios while remaining practical for real-world applications. The source code of the algorithm is available at \url{https://github.com/thuhangkhuat/PDE_target_search}.
}

\keywords{Optimal search, unmanned aerial vehicle, differential evolution, swarm intelligence}

\maketitle

\section{Introduction}

\begin{figure}
    \centering
    \includegraphics[width=\linewidth]{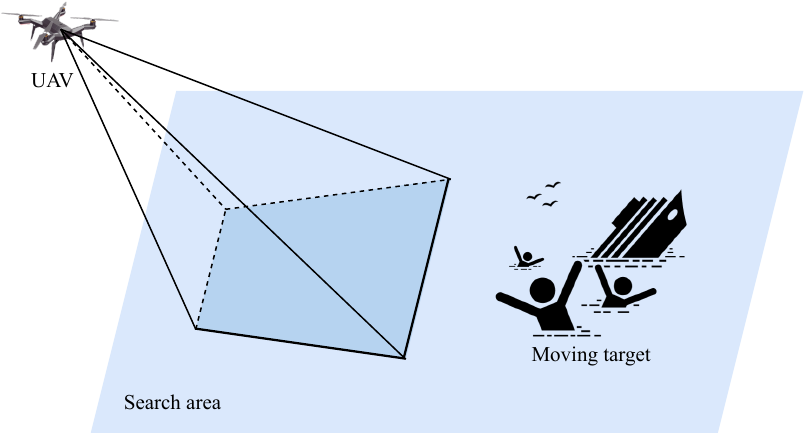}
    \caption{Moving target search problem}
    \label{fig:problem}
\end{figure}

Search and rescue operations involve locating and assisting missing persons in open environments, such as disaster zones, wilderness areas, or urban settings. Traditional methods often rely on ground-based teams and manned aircraft, which have limitations in flexibility and safety. Recently, unmanned aerial vehicles (UAVs) have emerged as a new means to overcome those limitations due to their ability to access remote and hazardous areas, gather real-time data, and provide aerial surveillance\mbox{~\cite{BOULARES2021103673,lee2023motion,8756125,9691840,phung2017enhanced}}. Figure {\ref{fig:problem}} illustrates the use of UAVs to search for moving targets. As time progresses, the probability of the target being found becomes smaller due to the degradation in initial information and the effect of external variables such as environmental conditions\mbox{~\cite{kamari2025wind,icsilak2026multi}}, terrain characteristics and target movement. Hence, the goal of using UAVs to look for a lost target involves determining a flight path that has the highest probability of finding the target given initial information on the target's location and search conditions~\cite{Foraker2015,Bourgault2006}.

To address the optimal search problem, probabilistic approaches that incorporate uncertainties associated with initial search conditions, belief map generation, and sensor models~\cite{1642081,Raap2016} are often used. They rely on the Bayesian framework to formulate objective functions, which are then used to evaluate the optimality of the solution~\cite{PHUNG2020106705,PEREZCARABAZA2018789}. To represent the initial search map, a multivariate normal distribution with parameters calculated based on the available information on the target's location is used~\cite{PHUNG2020106705,MahmoudZadeh2018}. Since the number of feasible solutions grows exponentially with respect to the search time and space \cite{Bernstein2002}, the resulting search problem is NP-hard. Consequently, obtaining an exact solution using classical optimization techniques such as differential calculus is impractical. Instead, approximation and metaheuristic methods including greedy algorithms \cite{DELAVERNHE2021469}, gradient-based methods \cite{5979704}, ant colony optimization (ACO) \cite{PEREZCARABAZA2018789}, particle swarm optimization (PSO) \cite{PHUNG2020106705}, and Bayesian optimization approaches (BOA) \cite{10.1145/2463372.2463417} are commonly employed.

At a high level, compared with other contemporary paradigms (e.g., deterministic optimization or learning-based approaches), metaheuristic methods offer a favorable balance between computational efficiency, flexibility in integrating complex constraints, and applicability under limited prior information. Table~\mbox{\ref{tbl:overview}} summarizes and compares the key characteristics of representative metaheuristic methods as discussed below.

\begin{table*}
\centering
\caption{Comparison among the search methods}
\label{tbl:overview}
\begin{tabular}{ccccccc}
\hline
Work       & Method  & Target        & Sensor     & Env.         & Constrained & Search model  \\ \hline
\cite{PHUNG2020106705} & PSO    & Dynamic       & Binary     & D            & \checkmark     & MDP     \\
\cite{PEREZCARABAZA2018789} & ACO            & Dynamic       & Binary     & D            & \checkmark  &mTS        \\
\cite{doi:10.1137/1019060} & Lagrange   & Static        & Stochastic & D/C &   &mTS          \\
\cite{10.1145/2463372.2463417} & BOA     & Dynamic       & Binary     & D            & \checkmark  &mTS         \\
\cite{5979704} & Gradient descent   & Static        & Stochastic & D          &   & MDP       \\
\cite{DELAVERNHE2021469} & Greedy            & Dynamic       & Binary     & D            & \checkmark    & MDP       \\
Ours       & DE                & Dynamic       & Stochastic & C          & \checkmark      & MDP      \\ \hline
\end{tabular}
\end{table*}

The \textbf{target} can be static or dynamic. In~\cite{5979704}, the static model is used to simplify the planning process. On the other hand, the target is considered dynamic in~\cite{DELAVERNHE2021469,PEREZCARABAZA2018789,10.1145/2463372.2463417,PHUNG2020106705} and modeled as a Markovian process. This model is more practical due to the moving nature of targets, e.g., human, and the influence of environmental factors such as wind or ocean waves.

The \textbf{sensor} can be modeled in different ways depending on the actual sensor used. In~\cite{doi:10.1137/1019060,5979704}, the range model, where the detection likelihood is a function of the distance between the sensor and target positions, is used. Other works focus on vision sensors with binary outputs~\cite{DELAVERNHE2021469,PEREZCARABAZA2018789,10.1145/2463372.2463417,PHUNG2020106705}. Vision sensors provide rich information of the search area but are affected by external factors such as lighting conditions and noise.  Those factors should be taken into account when developing the target detection model.

The \textbf{environment} can be modeled as a continuous world with piece-wise linear control actions or a discrete world with a discrete set of actions. In~\cite{PHUNG2020106705,PEREZCARABAZA2018789,10.1145/2463372.2463417}, the search environment is represented as a grid-based discrete environment so that the movements of the target can be simplified to eight main directions. Studies in~\cite{doi:10.1137/1019060,5979704}, on the other hand, consider continuous space and model the target via transition probability functions. The use of continuous space is more generic and overcomes the limitations related to resolution and scalability. 

\textbf{Constraints} are often added to the target model to reduce uncertainty in their dynamics. In some works, the Markovian process is used to represent the movement of the target so that its next state only depends on its current state~\cite{1642081,PHUNG2020106705}. The works in~\cite{PHUNG2020106705,PEREZCARABAZA2018789} further constrains the target to a deterministic Markovian process such that its trajectory can be determined if the initial location is known. Other works limit the target's movement to certain directions through the modeling of external factors that influence its movement such as wind and ocean currents~\cite{Bourgault2006,10.1145/2463372.2463417}. Constraints are necessary for dynamic target search since they reduce the complexity of the search problem.


The \textbf{search model} is classified into two main categories, minimum time search (mTS)~\cite{PEREZCARABAZA2018789,doi:10.1137/1019060,10.1145/2463372.2463417} and maximum detection probability (MDP)~\cite{PHUNG2020106705,5979704,DELAVERNHE2021469}. While mTS and MDP are different, there is a correlation between them so that obtaining one goal also leads to a good result for the other goal.

Apart from the above algorithms, differential evolution (DE) is a metaheuristic algorithm widely used for solving optimization problems~\cite{2001,Ahmad2022}. It is particularly effective in continuous, multi-dimensional search spaces. One of the main advantages of DE is its computational efficiency in arithmetic operations to create new candidate solutions. DE also exhibits strong global search capabilities through the differential mutation process, enabling it to escape local optima and converge towards the global optimum~\cite{Deng2021,price2006differential}. Additionally, DE is not sensitive to parameter settings, making it more versatile compared to other optimization algorithms~\cite{Civicioglu2019,Piotrowski2023}.

In this study, we propose a new algorithm for the optimal search problem that exploits the DE algorithm in polar coordinates to maximize the probability of finding a moving target. The probability of finding the target is modeled based on the analysis of the characteristics of the visual sensor, its footprint and the human vision model. The proposed algorithm also considers the constraints imposed by the UAV's kinematics and the target's dynamics. The results from a number of comparisons and experiments show that the proposed algorithm outperforms other state-of-the-art algorithms in both detection probability and search time while remaining computationally efficient. Our contributions in this work are threefold: (i) formulating the search problem as an optimization problem that considers practical aspects of the search problem including the search space, target dynamics, sensor models and initially available information; (ii) developing a new polar coordinate-based DE algorithm (PDE) that incorporates kinematic constraints to narrow the search space and exploits maneuver characteristics of the UAV to mitigate local optima and achieve improved convergence; (iii) evaluating the practicability of the proposed method via experimental search scenarios with real UAVs.

The remaining sections of this paper are organized as follows. Section 2 describes the process of deriving an objective function for the visual search problem. Section 3 introduces the proposed polar coordinate-based Differential Evolution (PDE) algorithm. Section 4 presents comparison and experiment results. The paper ends with conclusions drawn in Section 5.

\section{Problem Formulation}
The problem of searching for a lost target is formulated via the modeling of the camera, moving target and belief map with details as follows.

\subsection{Moving target model}
Consider a target $t$ moving in the environment with its location represented by $\xi^t_k$. Initially, a belief map is used to model the probability of the target location based on the available information. This belief map is described by a Gaussian distribution centered at the target's last known location, $\xi^t_0$~\cite{1642081}. A location $s_k$ in the belief map then has the probability of being the target's location at time $k$ given by:

\begin{equation}
    p\left(s_k\right)=\dfrac{1}{2\pi\sigma_k^2}
    e^{-\dfrac{\left\|s_k-\xi_k^t\right\|^2}{2\sigma_k^2}},
\label{eqn:gauss}
\end{equation}
where $\sigma_k$ is the standard deviation of the Gaussian distribution. 

In moving target search, a Markov process is often used to model the target dynamics so that its present location $\xi^t_{k}$ only depends on its previous location, $\xi^t_{k-1}$~\cite{1642081,Bourgault2006,PHUNG2020106705}. In this work, we consider the target dynamics as a deterministic Markov process where the transition function, $p(\xi^t_k|\xi^t_{k-1})$, representing the probability that the target moves from $\xi^t_{k-1}$ to $\xi^t_k$, is known. As a result, the trajectory of target $\xi^t$ can be determined if its exact initial location is known. This assumption is reasonable in several real-world scenarios, such as a floating survivor at sea, where the target dynamics can be approximated as deterministic given known ocean conditions. In more general cases, uncertainty in the target's motion is partially captured by the Gaussian belief map defined in \mbox{\eqref{eqn:gauss}}, where the standard deviation $\sigma_k$ reflects uncertainty in the target's location. In this model, the uncertainty $\sigma_k$ is assumed to be constant over time according to the Steady
State framework \mbox{~\cite{thrun2002probabilistic,akhlaghi2017adaptive}}. Consequently, this modeling assumption has been widely adopted in prior studies \mbox{\cite{iida1998optimal, PEREZCARABAZA2018789, PEREZCARABAZA2019357, raap2017aerial}}.

\subsection{Target detection model}



\begin{figure*}
\centering
    \begin{subfigure}[b]{0.35\textwidth}
    \centering
    \includegraphics[width=\textwidth]{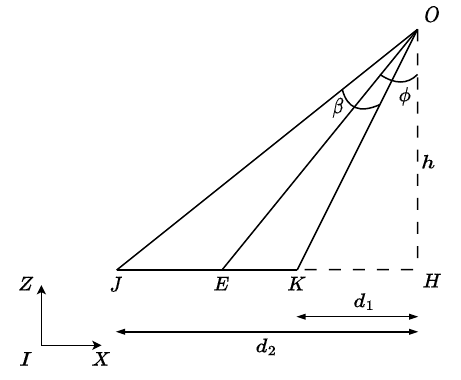}
    \caption{Side view}
    \label{fig:view_side}
    \end{subfigure}
    \hspace{2cm}
    \begin{subfigure}[b]{0.3\textwidth}
    \centering
    \includegraphics[width=\textwidth]{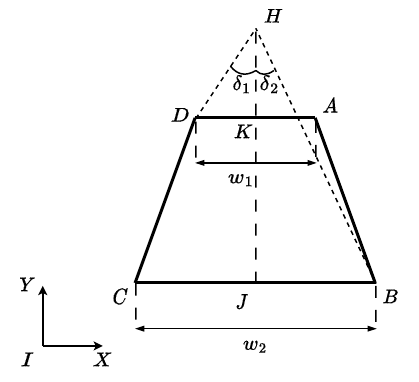}
    \caption{Top view}
    \label{fig:view_top}
    \end{subfigure}
    \begin{subfigure}[b]{0.5\textwidth}
    \centering
    \includegraphics[width=\textwidth]{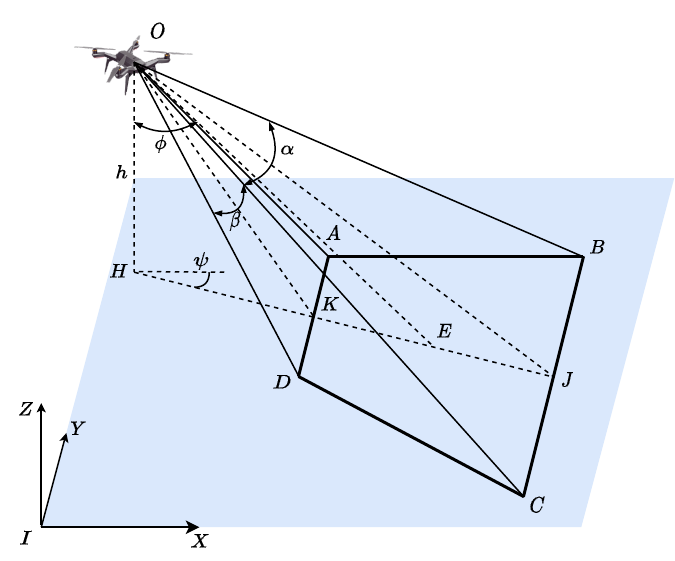}
    \caption{3D view}
    \label{fig:view_3d}
    \end{subfigure}
    \caption{The camera footprint in global coordination $IXYZ$}
    \label{fig:camerafp}
\end{figure*}

In this work, an RGB camera is used to detect the target. The camera is attached to the center $O(x,y,z)$ of the UAV via a three-axis gimbal. The gimbal controls the camera to point downward at a fixed angle $\phi$. Denote $\alpha$ and $\beta$ respectively as the horizontal and vertical view angles of the camera, $\psi$ as the heading angle of the UAV, and $h$ as the flight altitude. The footprint of the camera then can be computed as illustrated in Figure \ref{fig:camerafp}. Specifically, the camera's looking angle $\phi$ is equal to $\angle EOH$, where $H$ is the projection of the UAV's center on the ground plane, and $E$ is the center of the camera's footprint, i.e., $E=AC\cap BD$. Denote $A$, $B$, $C$, and $D$ respectively as four vertices of the camera's footprint representing the detection area. Denote $J$ and $K$ as the midpoints of $AD$ and $BC$, respectively. According to Figure \ref{fig:view_side}, distances from $H$ to each parallel side of the trapezoid $ABCD$ are computed as:
\begin{equation}
\begin{aligned}
    d_1 &= h\tan\left(\phi-\beta/2\right) \\
    d_2 &= h\tan\left(\phi+\beta/2\right)
\end{aligned}
\end{equation}
The length of the parallel sides, $AD$ and $BC$, are then given by:
\begin{equation}
\begin{aligned}
    w_1 &= \dfrac{2h\tan\left(\alpha/2\right)}{\cos\left(\phi-\beta/2\right)}, \\
    w_2 &= \dfrac{2h\tan\left(\alpha/2\right)}{\cos\left(\phi+\beta/2\right)}
\end{aligned}
\end{equation}
Finally, four vertices, $A$, $B$, $C$, and $D$ are obtained as follows:
\begin{equation}
\begin{aligned}
    &x_A = x + \frac{w_1\cos(\psi-\delta_1)}{2\sin{\delta_1}}, y_A = y + \frac{w_1\sin(\psi-\delta_1)}{2\sin\delta_1}; \\
    &x_B = x + \frac{w_2\cos(\psi-\delta_2)}{2\sin\delta_2}, y_B = y + \frac{w_2\sin(\psi-\delta_2)}{2\sin\delta_2}; \\
    &x_C = x + \frac{w_2\cos(\psi+\delta_2)}{2\sin\delta_2}, y_C = y + \frac{w_2\sin(\psi+\delta_2)}{2\sin\delta_2}; \\
    &x_D = x + \frac{w_1\cos(\psi+\delta_1)}{2\sin{\delta_1}}, y_D = y + \frac{w_1\sin(\psi+\delta_1)}{2\sin\delta_1};
\end{aligned}
\label{eqn:ABCD}
\end{equation}
where $\delta_1$ and $\delta_2$ are the two angles shown in Figure \ref{fig:view_top} and computed as follows:
\begin{equation}
\begin{aligned}
    \delta_1 &= \tan^{-1}\left(\dfrac{w_1}{2d_1}\right)\\
    \delta_2 &= \tan^{-1}\left(\dfrac{w_2}{2d_2}\right).
\end{aligned}
\end{equation}
In the algorithmic implementation, these angles are evaluated using the two-argument arctangent function, $\text{atan2}(\cdot, \cdot)$, to ensure numerical stability.

During operation, the image obtained from the camera is transmitted to the base station, where an observer analyzes it to identify the target. Denote $p\left(D_k|\xi^t_k\right)$ as the probability that the observer finds the target at time $k$. This probability can be estimated using NVESD~\cite{hixson2004target}, a reliable model to evaluate the probability of target observation. This model works on the principle that the target detection performance depends on the image quality and the observer's vision. Denote TTP as the target tracking performance. It relates to the image uniformity and the observer's viewing threshold as follows~\cite{hixson2004target}:
\begin{equation} \label{eq:TTP}
\text{TTP}=\intop_{f_\text{min}}^{f_\text{max}}\sqrt{\dfrac{C_\text{tgt}}{\text{CTF}(f)}}df,
\end{equation}
where $f$ is the spatial frequency, CTF is the contrast threshold function, and  $C_\text{tgt}$ is the target contrast. CTF is the quantification of the observer's vision defined as in~\cite{hixson2004target}. The target contrast ($C_\text{tgt}$) represents the ratio of the mean and standard deviation of the target to the mean of the scene in the image obtained from the camera on the UAV~\cite{vollmerhausen2004new}. 
By applying the NVESD model, the likelihood of target detection is given as follows:
\begin{equation}
    p\left(D_k|\xi^t_k\right) = \dfrac{\left(\dfrac{V_k}{V_{50}}\right)^\gamma}{1+\left(\dfrac{V_k}{V_{50}}\right)^\gamma},
    \label{eq:detection}
\end{equation}
where
\begin{equation} \label{eq:gamma}
    \gamma = 1.51+0.24\left(\frac{V_k}{V_{50}}\right),
\end{equation}
\begin{equation}
    V_k = \text{TTP}\sqrt{\dfrac{S_{k}^{t}\cos^2\phi\tan\dfrac{\alpha}{2}g(\phi,\beta)}{S_{k}^\text{fp}}},
\end{equation}

\begin{equation}
\begin{aligned}
    g(\phi,\beta)=&\left[\tan\left(\phi+\beta/2\right)-\tan\left(\phi-\beta/2\right)\right]\times\\
    &\left[\dfrac{1}{\cos\left(\phi+\beta/2\right)}-\dfrac{1}{\cos\left(\phi-\beta/2\right)}\right],
\end{aligned}
\end{equation}
with $V_{50}$ is the critical period so that the probability of detecting the target is 0.5~\cite{espinola2007modeling}, $S^t_k$ is the area of the target, and $S_{k}^\text{fp}=S_\text{ABCD}$ is the camera footprint. The coefficients 1.51 and 0.24 in \mbox{\eqref{eq:gamma}} are empirical constants of the NVESD model obtained from regression fitting to human observer detection data, where 1.51 defines the baseline slope of the detection curve and 0.24 accounts for the increased steepness as target visibility improves. From (\ref{eq:detection}), the likelihood of not detecting the target at time $k$ is given by:
\begin{equation}
    p\left(\bar{D}_k|\xi^t_k\right) = 1-p\left(D_k|\xi^t_k\right).
    \label{eqn:actual}
\end{equation}

\subsection{Estimation of the target's location}
The estimation of the target's location is carried out recursively via two phases, prediction and update, based on the Bayesian theory as follows.

\subsubsection{Prediction phase}
Denote $p(\xi^t_{k}|\zeta_{1:k-1})$ as the probability that the target is located at $\xi^t_{k}$ given all observations from $\zeta_1$ to $\zeta_{k-1}$. As the estimation process is recursive with the initial belief map defined by \mbox{\eqref{eqn:gauss}}, we assume that at time $k$, $p(\xi^t_{k-1}|\zeta_{1:k-1})$ from the previous time step is available. 

The predicted location of the target then can be obtained from the following Chapman-Kolmogorov equation~\cite{Bourgault2006}:
\begin{equation}
    p(\xi^t_k|\zeta_{1:k-1}) = \int_S p(\xi^t_k|\xi^t_{k-1})p(\xi^t_{k-1}|\zeta_{1:k-1})d\xi^t_{k-1},
    \label{eqn:predict}
\end{equation}
where $p(\xi^t_k|\xi^t_{k-1})$ is the transition probability obtained from the Markov process representing the target's motion. 

For a deterministic process without noise, the target location $\xi^t_k = [x^t_k, y^t_k]^T$ is updated via the following kinematic function:

\begin{equation}
\begin{aligned}
    x^t_k &= x^t_{k-1} + v_{k-1} \Delta t \cos{\theta_{k-1}} \\
    y^t_k &= y^t_{k-1} + v_{k-1} \Delta t \sin{\theta_{k-1}}, 
\end{aligned}
\end{equation}
where $v_{k-1}$ is the speed, $\theta_{k-1}$ is the heading angle, and $\Delta t$ is the time step interval. Let $\mathbf{v}_{k-1} = [v_{k-1} \cos\theta_{k-1}, v_{k-1} \sin\theta_{k-1}]^T$ represent the velocity vector. The transition probability $p(\xi^t_k|\xi^t_{k-1})$ is then represented by a Dirac delta function $\delta$ rather
than a Gaussian distribution. The probability is $1$ if the target moves exactly according to the kinematic equation and $0$ everywhere else. Therefore, the practical form of the transition function is expressed as:
\begin{equation}
    p(\xi^t_k \mid \xi^t_{k-1}) = \delta \left( \xi^t_k - (\xi^t_{k-1} + \mathbf{v}_{k-1} \Delta t) \right).
\end{equation}




\subsubsection{Update phase} \label{sec:update_phase}
The updated location of the target at time $k$ is computed based on the Bayes rule when observation $\zeta_k$ becomes available. It is performed by multiplying the predicted location in (\ref{eqn:predict}) by the new conditional observation likelihood $p(\zeta_k|\xi^t_k)$ as follows:
\begin{equation}
    p(\xi^t_k|\zeta_{1:k})=Kp(\xi^t_k|\zeta_{1:k-1})p(\zeta_k|\xi^t_k),
\end{equation}
where $K$ is the normalization factor given by:
\begin{equation}
    K=1/\int{p(\xi^t_k|\zeta_{1:k-1})p(\zeta_k|\xi^t_k)}d\xi^t_k.
    \label{eqn:K}
\end{equation}

\subsection{Visual search fitness function}
Let $r_k = p(\bar{D}_k|\zeta_{1:k-1})$ be the probability that the target does not get detected at time $k$. According to the Bayesian theory, $r_k$ depends on two factors: (i) the likelihood of no detection described in \eqref{eqn:actual}, and (ii) the latest prediction of the target's location expressed in  \eqref{eqn:predict}. That probability is given by:
\begin{equation}
    r_k=\int_S{p(\bar{D}_k|\xi^t_k)p(\xi^t_k|\zeta_{1:k-1})d\xi^t_k},
\end{equation}
where $S$ is the searching area. Observe that when there is a ``no detection'' event ($\zeta_k = \bar{D}_k$), the value of $r_k$ becomes precisely the inverse of the normalization factor $K$ from \eqref{eqn:K}, i.e., $r_k=1/K$, and consequently, it is less than 1. Let $R_k=p(\bar{D}_{1:k})$ be the joint probability of not detecting the target from time 1 to $k$. We can obtain $R_k$ by multiplying the probability that the target does not get detected at each time step over time:
\begin{equation}
    R_{k}=\prod_{i=1}^{k}r_{i}=R_{k-1}r_{k}.
\end{equation}

Let $p_{k}$ be the probability that the target gets detected at time $k$. It can be calculated as:
\begin{equation}
    p_{k}=\prod_{i=1}^{k-1}r_{i}\left(1-r_{k}\right)=R_{k-1}\left(1-r_{k}\right)
    \label{eqn:detectProb}
\end{equation}

The cumulative probability that the target gets detected in $k$ steps can be computed by adding $p_k$ over $k$ steps:
\begin{equation}
    P_{k}=\sum_{i=1}^{k}p_{i}=P_{k-1}+p_{k}
    \label{eqn:Pk}
\end{equation}
It can be seen from (\ref{eqn:detectProb}) and (\ref{eqn:Pk}) that the probability of detecting the target for the first time $p_k$ decreases over time because the chance that the target gets detected in previous steps becomes larger. The cumulative probability $P_k$ is therefore bounded and goes toward one as time $k$ tends to infinity. 

In practice, the search time is finite, i.e., $k \in \{1,\cdots,N\}$, where $N$ is the maximum search time. The objective $J$ of the search strategy is therefore to find a flight path that maximizes the cumulative probability $P_N$ within that given search time. The fitness function for the search problem then can be defined as:
\begin{equation}
    J= P_N = \sum_{k=1}^Np_k.
    \label{eqn:cost}
\end{equation}

Equation \eqref{eqn:cost} turns the search problem into an NP-hard optimization problem where the number of possible solutions increases exponentially with respect to the search time and space~\cite{Bernstein2002}. Hence, traditional methods to find the exact solution such as differential calculus cannot solve it in polynomial time. Instead, meta-heuristic algorithms are often used~\cite{Rajwar2023}.

\section{Polar coordinate-based Differential Evolution algorithm}

Among meta-heuristic search algorithms, DE is one of the most popular algorithms with the capability to solve challenging optimization problems~\cite{8937719,Civicioglu2019,Ahmad2022} while maintaining computational efficiency~\cite{Das2011}. This study proposes an improved DE algorithm for moving target search with details as follows.

\subsection{Standard differential evolution algorithm}
The differential evolution (DE) algorithm is a population-based optimization method that belongs to the class of evolutionary algorithms~\cite{price2006differential}. It operates based on three computational operators named \textit{mutation},
\textit{crossover} and \textit{selection}.

The \textbf{mutation operator} forms new candidate solutions by adding the difference between two individuals to an existing individual. For each individual $u_{ij}$ in the $j$-th generation, a mutant solution $x_{ij}$ is generated as:
\begin{equation}
    x_{ij} = u_{ij} + F \left(u_{n_1,j} - u_{n_2,j}\right),
\end{equation}
where $i\in\left\{1,2,\cdots,N_p\right\}$ is the individual index; $n_1$ and $n_2$ are random indexes selected from $\left\{1,2,\cdots,N_p\right\}$ so that $n_1 \neq n_2 \neq i$; $F>0$ is a weight factor. 

The \textbf{crossover operator} combines the mutant solutions with the existing solutions to form trial solutions as follows:

\begin{equation}
    o_{ij}=\begin{cases}
x_{ij} & \text{if }\text{rand}_{i}\left(0,1\right)\leq p_{Cr}\\
u_{ij} & \text{otherwise}
\end{cases}
\end{equation}
where $o_{ij}$ is a trial solution, $p_{Cr} \in [0,1]$ is 
 a predetermined crossover probability.

The \textbf{selection operator} selects better individuals between the trial and existing solutions based on a fitness function $f$. It is carried out as follows:
\begin{equation}
    u_{i,j+1}=\begin{cases}
o_{ij} & \text{if }f\left(o_{ij}\right)>f\left(u_{ij}\right)\\
u_{ij} & \text{otherwise}
\end{cases}
\end{equation}

Those three operators enable the algorithm to improve its solution through each generation and gradually converge to the optimal one. However, the standard DE algorithm was originally developed for generic optimization problems with the objective of minimizing a cost function~\cite{price2006differential}. As a result, it does not exploit UAV-specific characteristics to improve exploration of the solution space. More critically, it fails to account for UAV kinematic and physical constraints, such as turning angles and speed limits, resulting in infeasible paths that the UAV cannot follow to find its target. To overcome these limitations, we propose a polar-based DE algorithm designed specifically for UAV path planning.

\subsection{The polar coordinate-based DE algorithm}
When using the DE algorithm for the search problem, each individual in the population represents a candidate solution, which is a flight path. This flight path is described by a set of waypoints or nodes that the UAV needs to travel through. Let $\xi_k = \left[x_k,y_k,z_k\right]^T$ be waypoint $k$. A flight path $\xi$ with $N$ waypoints can be represented as:

\begin{equation}
    \xi = \left(x_1,y_1,z_1,...,x_N,y_N,z_N\right)
\end{equation}

As the UAV movement is characterized by its kinematic parameters such as yaw angle, using Cartesian coordinates makes the algorithm unable to deal with the UAV's related constraints due to the lack of a direct relationship between the Cartesian coordinates and kinematic parameters. In this work, we propose to use polar coordinates to overcome those limitations. Consider the discrete-time UAV kinematics as follows:
\begin{equation}
    \begin{aligned}
        x_{k+1}&= x_k + \rho_k\cos\theta_k\cos\psi_k\\
        y_{k+1}&= y_k + \rho_k\cos\theta_k\sin\psi_k\\
        z_{k+1}&= z_k + \rho_k\sin\theta_k,
    \end{aligned}
\end{equation}
where $\rho_k$ is the step length between waypoint $k$ and $k+1$, and $\theta_k$ and $\psi_k$ are respectively the pitch and yaw angles of the UAV. Since the UAV flies at a constant altitude during operation~\cite{Alcntara2021}, the climbing angle $\theta_k$ is negligible. The step length $\rho_k$ and yaw angle $\psi_k$ are constrained to the range $\rho_\text{min}\leq\rho_k\leq\rho_\text{max}$ and $\psi_\text{min}\leq\psi_k\leq\psi_\text{max}$ due to physical limits such as velocities of the UAV. We incorporate those constraints via the use of polar coordinates. Specifically, we encode each path as a set of line segments. Each segment represents the movement of the UAV from one waypoint to another and thus can be described by a vector. The vector is represented in the polar coordinate system with two elements, magnitude $\rho_k\in\left(0,\rho_\text{max}\right)$ and angle $\psi_k\in\left(-\pi,\pi\right)$. A flight path $U$ in the polar coordinate system is represented as:
\begin{equation}
    U = \left(\rho_1,\psi_1,...,\rho_k,\psi_k,...,\rho_N,\psi_N\right).
\end{equation}



The mutation of the PDE algorithm then can be implemented as follows:
\begin{equation}\label{equ:mutation}
     X_{ij} = U_{ij} + F (U_{n_1,j} - U_{n_2,j}),
\end{equation}
where $X_{ij}$ is a mutant individual, $i$ is the individual index, and $j$ is the generation index.  The crossover operator $O_{ij}$ of the PDE can be obtained as follows:
\begin{equation}
    O_{ij}=\begin{cases}
X_{ij} & \text{if }\text{rand}_{i}\left(0,1\right)\leq p_{Cr}\\
U_{ij} & \text{otherwise}
\end{cases}
\label{equ:crossover}
\end{equation}

To calculate the value of the fitness function $f$ in \eqref{eqn:cost}, it is necessary to convert a flight path $U$ in the polar coordinate system to its correspondence $\xi$ in the Cartesian coordinate. Each waypoint in the path can be converted as follows:
\begin{equation}
    \xi_{k+1} = \xi_k + \left[\begin{array}{c}
\rho_{k}\cos\psi_{k}\\
\rho_{k}\sin\psi_{k}\\
0
\end{array}\right]=\left[\begin{array}{c}
x_{k+1}\\
y_{k+1}\\
h
\end{array}\right]
    \label{equ:PolartoCart}
\end{equation}

Denote the mapping in (\ref{equ:PolartoCart}) as $f:U\to\xi$. The selection operator $U_{i,j+1}$ then can be expressed through the fitness function in \eqref{eqn:cost} as:
\begin{equation}
    U_{i,j+1}=\begin{cases}
O_{ij} & \text{if }J\left(f(O_{ij})\right)>J\left(f(U_{ij})\right)\\
U_{ij} & \text{otherwise}
\end{cases}
\label{equ:select}
\end{equation}

In this new implementation, constraints on the kinematics of the UAV can be directly integrated into the PDE by limiting the range of polar variables, i.e., $\rho_\text{min}\leq\rho_k\leq\rho_\text{max}$ and $\psi_\text{min}\leq\psi_k\leq\psi_\text{max}$. As a result, the search space is narrowed down to give a higher probability of finding the optimal solution in a shorter time. The direct relationship between the magnitude and angle components of polar coordinates and the velocity and yaw angle of the UAV enables the PDE to search for solutions in the configuration space to achieve better results. Most importantly, the integration of kinematic constraints and physical limits into the PDE ensures that the generated paths are feasible for the UAV to follow.


\begin{algorithm}
\caption{Pseudocode of the PDE algorithm}\label{alg:pseudo}
    \tcc{Initialization}
    Get target dynamics and initial location data\;
    Create belief map\;
    Initialize population parameters $N_p$, $F$, $p_{Cr}$\;
    \ForEach {individual $i$ in population}{
        Create random polar coordinate paths $U_i$\;
    }
    \tcc{Evolutions}
    \For{$j \leftarrow 1$ to max iterations}
    {
        \ForEach{ individual $i$ in population }
        {
            Select $n_1, n_2 \in [1,...,N_p]$, $n_1 \neq n_2 \neq i$\;
            Create $X_{ij}$ using mutation operator\tcc*[r]{Mutation - Eq.\ref{equ:mutation}}
            Create $O_{ij}$ using crossover operator\tcc*[r]{Crossover - Eq.\ref{equ:crossover}}
            Convert $O_{ij} \rightarrow \xi_{ij}$\tcc*[r]{Eq.\ref{equ:PolartoCart}}
            Update fitness value of $\xi_{ij}$\tcc*[r]{Eq.\ref{eqn:cost}}
            Evaluate $U_{i,j+1}$ and perform selection operation\tcc*[r]{Selection - Eq.\ref{equ:select}}
            }
        Update maximum cost value $J_\text{max}$\;
    }
\end{algorithm}

The implementation of the PDE together with the equations used are presented in Algorithm \ref{alg:pseudo}. It has a similar structure to the standard DE such as parameter initialization, individual generation, and population evolution. It, however, differs in the representation of individuals, mapping from the polar to Cartesian coordinates, and evaluating equations.

\section{Results}
To evaluate the performance of the proposed PDE algorithm, a number of simulations, comparisons and experiments have been conducted with details described as follows.

\subsection{Scenarios setup}
Since the detection probability accumulates as the UAV scans the search area, it is proportional to the path length and the UAV's execution time. These factors vary depending on the type of UAV used, such as fixed-wing or quadcopter, and its capabilities, including battery life and flight speed. We account for these variations by evaluating the algorithm across multiple scenarios with different map sizes,  and numbers of search steps. Specifically, the evaluation scenarios are designed with parameters as follows:
\begin{enumerate}[label=\roman*.]
    \item The search map has the size from $500 \text{ m} \times 500 \text{ m}$ to $2000 \text{ m} \times 2000 \text{ m}$ to measure the search capacity of the algorithm, especially when dealing with large search areas.
    \item The number of search steps is chosen as 30, 60, 100, and 200 nodes, respectively, to evaluate the scalability and processing time of the algorithm.
    \item The initial position and moving direction of the target are set to different values to evaluate the adaptability of the algorithm.
\end{enumerate}

\begin{figure*}
\centering
    \begin{subfigure}[b]{0.49\textwidth}
    \centering
    \includegraphics[width=\textwidth]{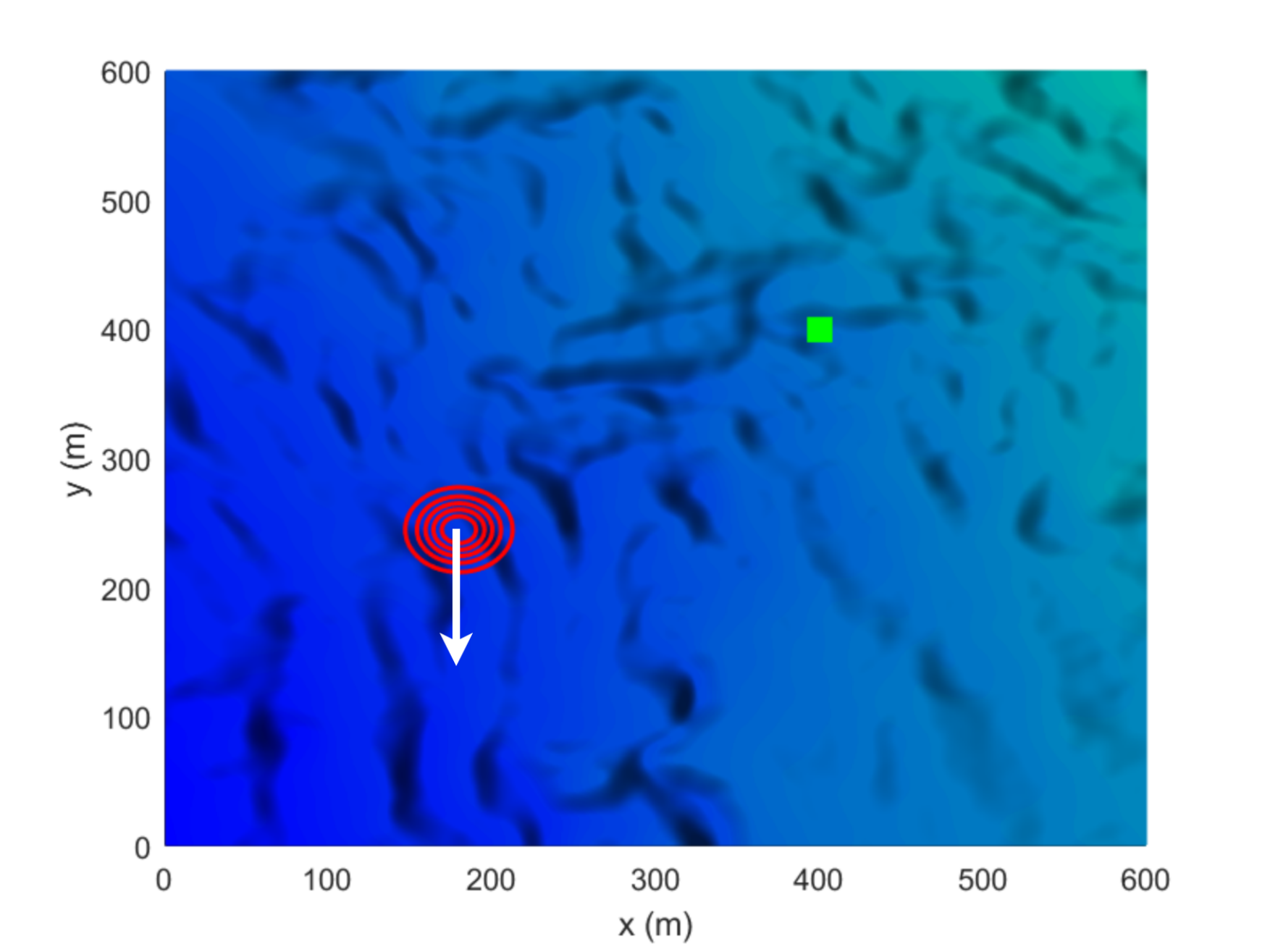}
    \caption{Scenario 1 - $600 \text{ m} \times 600 \text{ m}$}
    \label{fig:scen1_init}
    \end{subfigure}
    \begin{subfigure}[b]{0.49\textwidth}
    \centering
    \includegraphics[width=\textwidth]{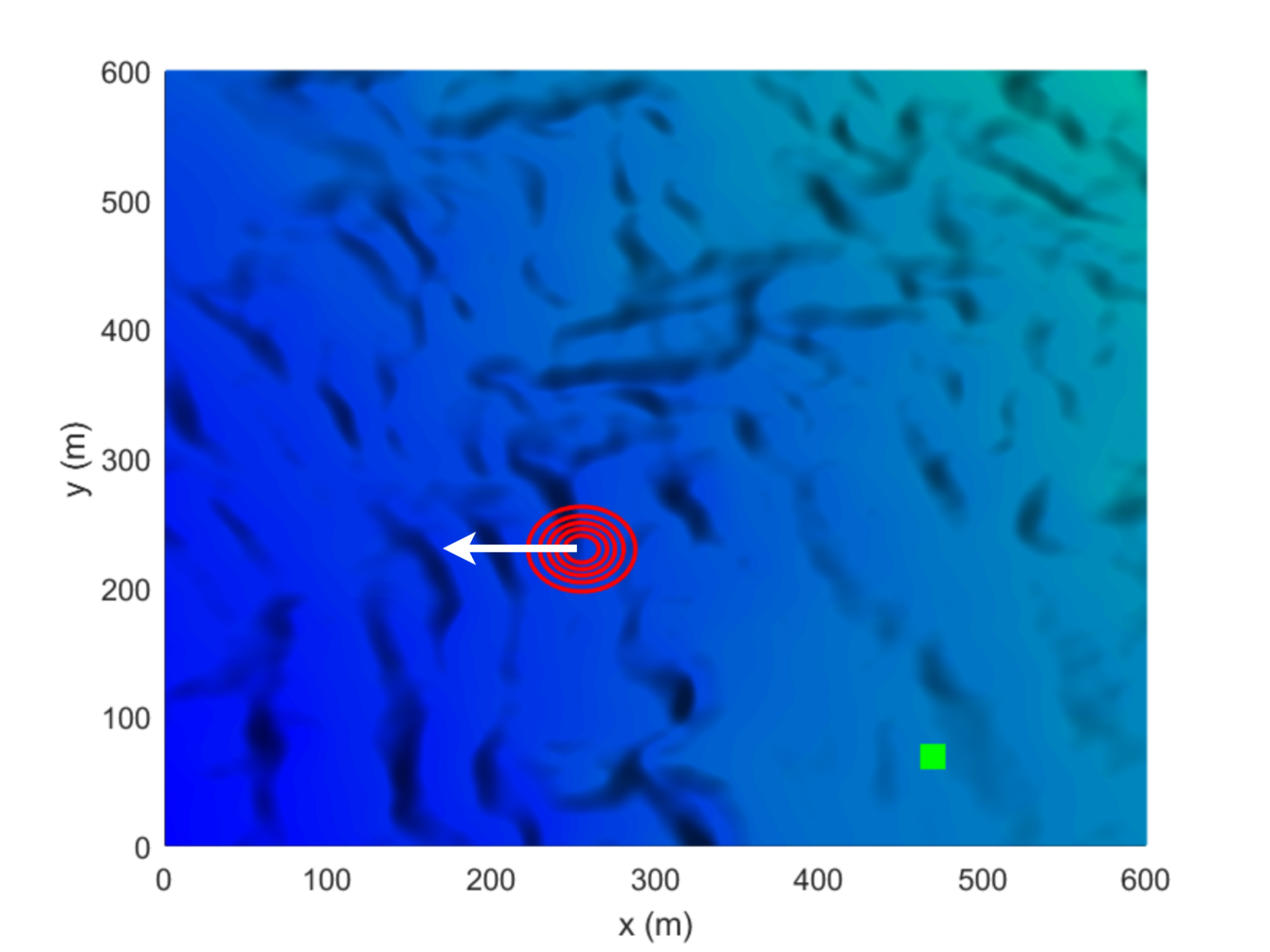}
    \caption{Scenario 2 - $600 \text{ m} \times 600 \text{ m}$}
    \label{fig:scen2_init}
    \end{subfigure}
    \begin{subfigure}[b]{0.49\textwidth}
    \centering
    \includegraphics[width=\textwidth]{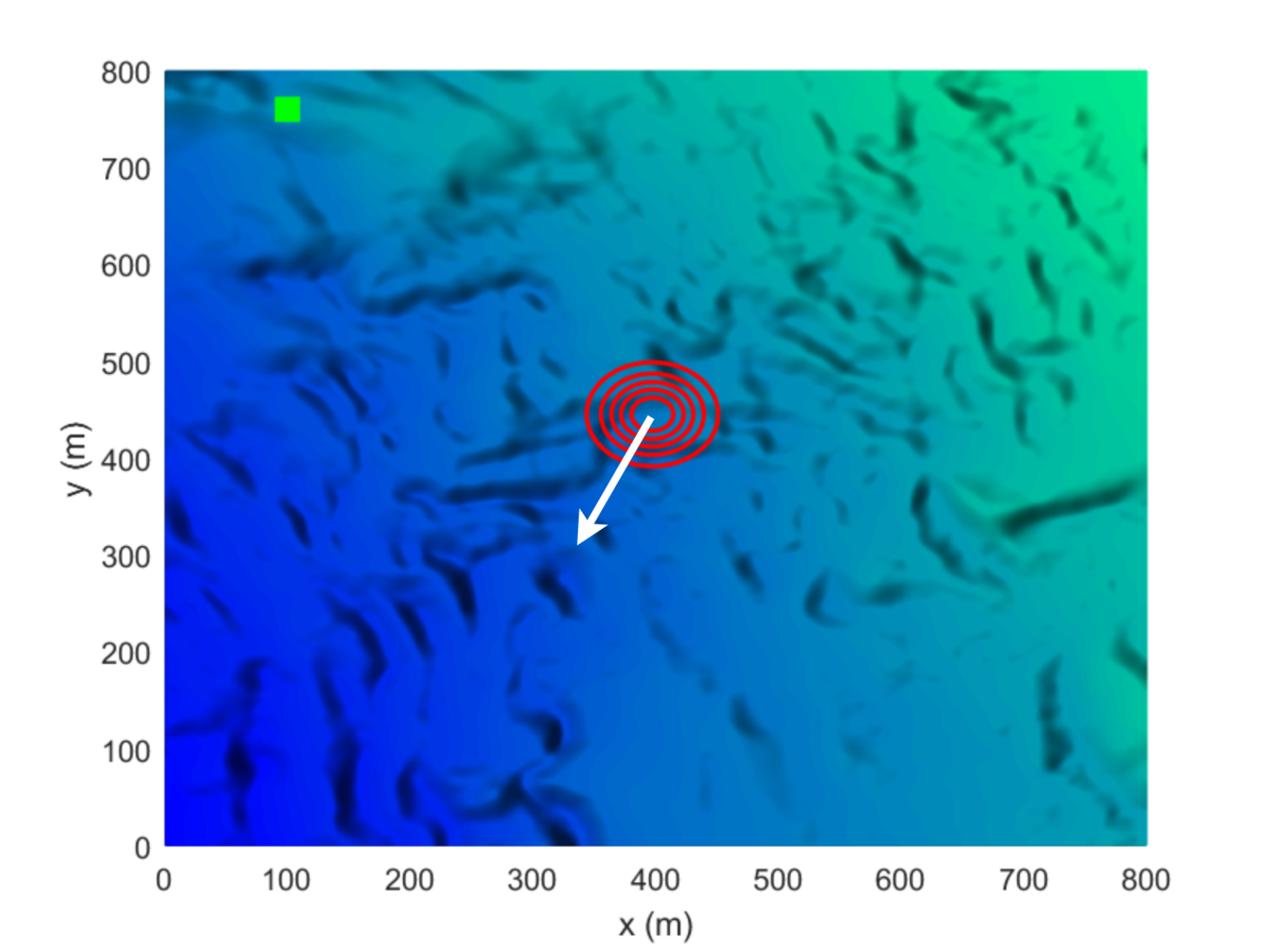}
    \caption{Scenario 3 - $800 \text{ m} \times 800 \text{ m}$}
    \label{fig:scen3_init}
    \end{subfigure}
    \begin{subfigure}[b]{0.49\textwidth}
    \centering
    \includegraphics[width=\textwidth]{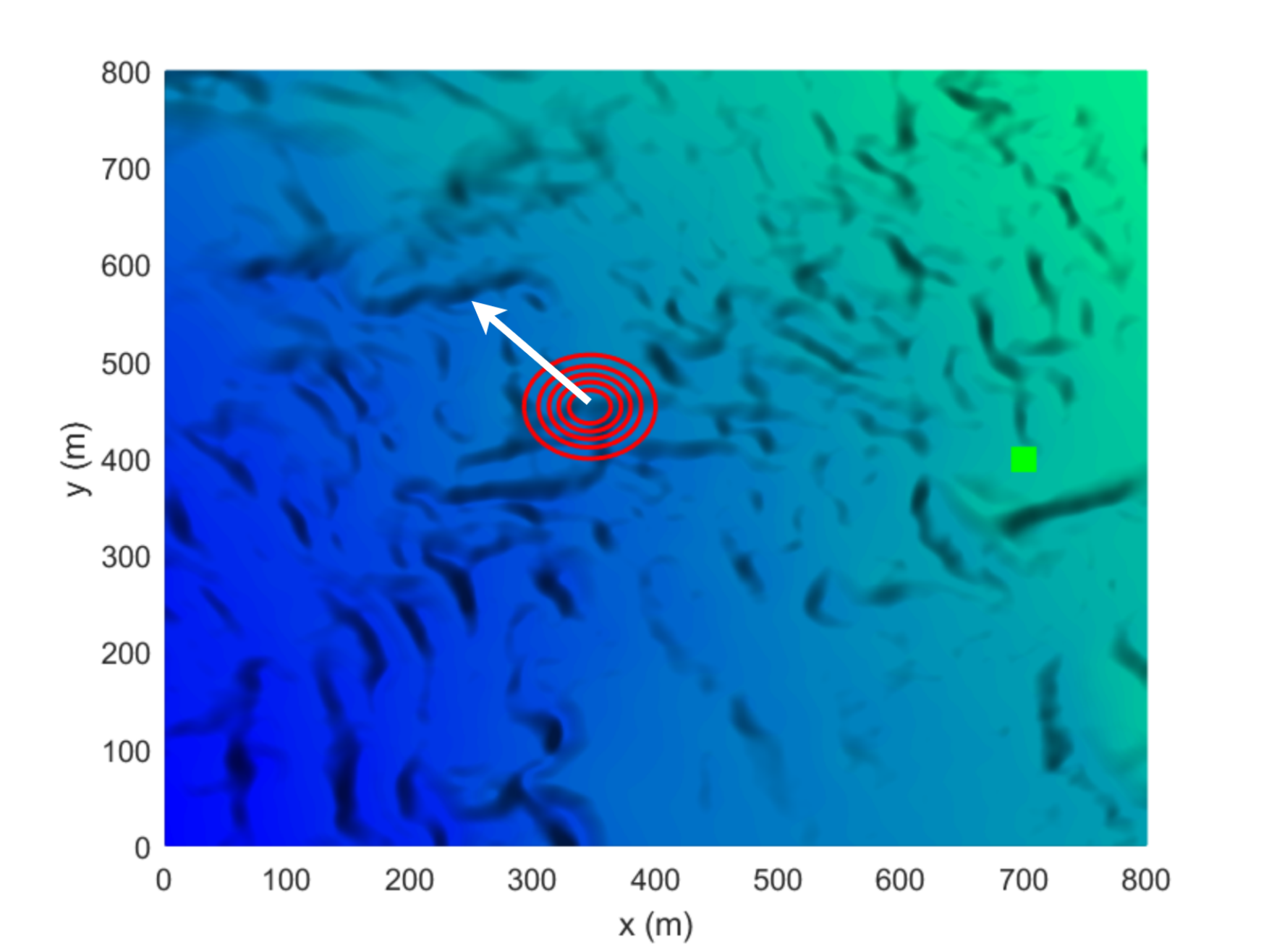}
    \caption{Scenario 4 - $800 \text{ m} \times 800 \text{ m}$}
    \label{fig:scen4_init}
    \end{subfigure}
    \begin{subfigure}[b]{0.49\textwidth}
    \centering
    \includegraphics[width=\textwidth]{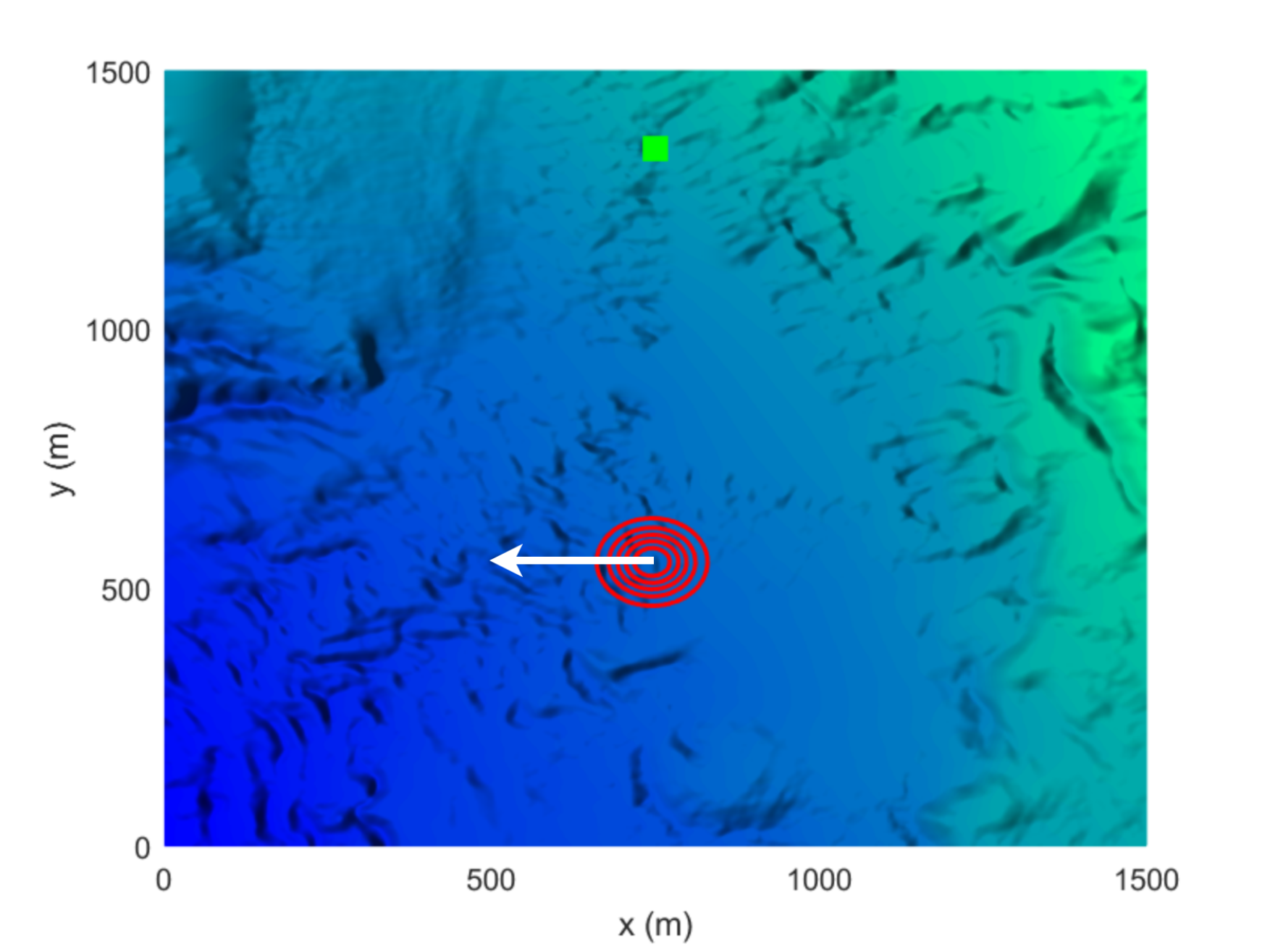}
    \caption{Scenario 5 - $1500 \text{ m} \times 1500 \text{ m}$}
    \label{fig:scen5_init}
    \end{subfigure}
    \begin{subfigure}[b]{0.49\textwidth}
    \centering
    \includegraphics[width=\textwidth]{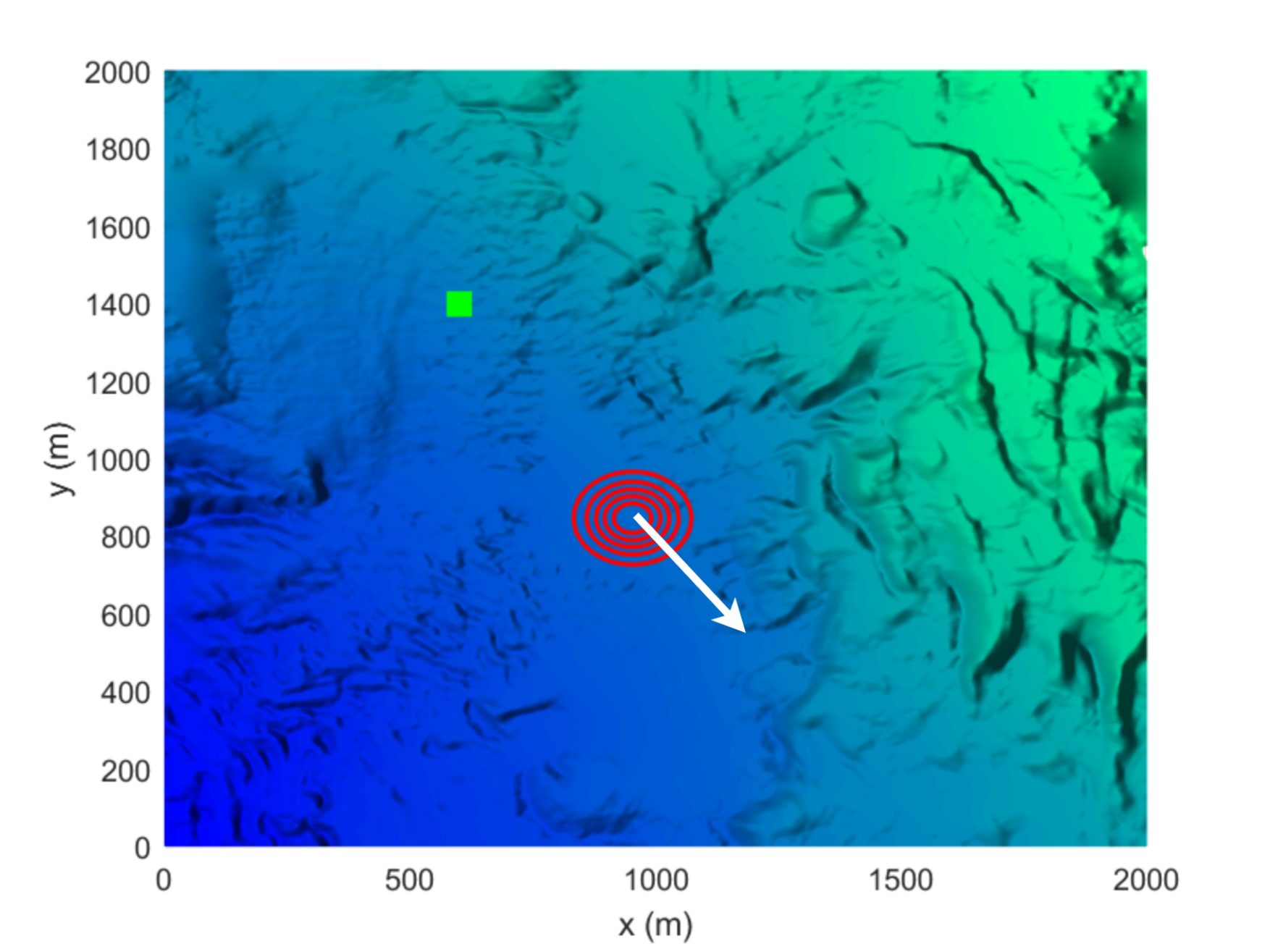}
    \caption{Scenario 6 - $2000 \text{ m} \times 2000 \text{ m}$}
    \label{fig:scen6_init}
    \end{subfigure}
    \caption{Arrangement of the UAV location and target movement in evaluating scenarios}
    \label{fig:init_scen}
\end{figure*}

Six scenarios were created from the above parameters representing different search situations as shown in Figure \ref{fig:init_scen}, where the green square symbol indicates the initial location of the UAV, the red circle represents the center of the target's probability density function (PDF), and the white arrow represents the direction of the target's movement. Detailed parameters for each scenario are given as follows:

\textbf{Scenario 1} (Figure \ref{fig:scen1_init}): the map size is $600 \text{ m} \times 600 \text{ m}$, the search path includes 30 waypoints, the target moves in the $- \pi/2$ direction, and the flight altitude is 10~m.

\textbf{Scenario 2} (Figure \ref{fig:scen2_init}): the map size is $600 \text{ m} \times 600 \text{ m}$, the search path includes 30 waypoints, the target moves in the $\pi$ direction, and the flight altitude is 10~m.

\textbf{Scenario 3} (Figure \ref{fig:scen3_init}): the map size is $800 \text{ m} \times 800 \text{ m}$, the search path includes 60 waypoints, the target moves in the $- 2.7\pi/4$ direction, and the flight altitude is 10~m.

\textbf{Scenario 4} (Figure \ref{fig:scen4_init}): the map size is $800 \text{ m} \times 800 \text{ m}$, the search path includes 60 waypoints, the target moves in the $3\pi/4$ direction, and the flight altitude is 10~m.

\textbf{Scenario 5} (Figure \ref{fig:scen5_init}): the map size is $1500 \text{ m} \times 1500 \text{ m}$, the search path includes 100 waypoints, the target moves in the $\pi$ direction, and the flight altitude is 15~m.

\textbf{Scenario 6} (Figure \ref{fig:scen6_init}): the map size is $2000 \text{ m} \times 2000 \text{ m}$, the search path includes 200 waypoints, the target moves in the $\pi/4$ direction, and the flight altitude is 15~m.

In our evaluations, the scaling factor $F$ is randomly sampled from the interval $[0.1,0.4]$ at each iteration. This range is smaller than the commonly used interval $F\in[0,1]$ in standard DE\mbox{~\cite{price2006differential}} and was selected to provide better exploitation performance in practice. Other parameters of the PDE and the camera are chosen as follows: the crossover probability $p_{Cr}=0.9$; the population size $N_p = 500$; the number of iterations is 100; the camera's angles of view $\alpha=80^\circ$ and $\beta=60^\circ$; and the pointing angle $\phi=30^\circ$.

\subsection{Results}
\begin{figure*}
\centering
    \begin{subfigure}[b]{0.45\textwidth}
    \centering
    \includegraphics[width=\textwidth]{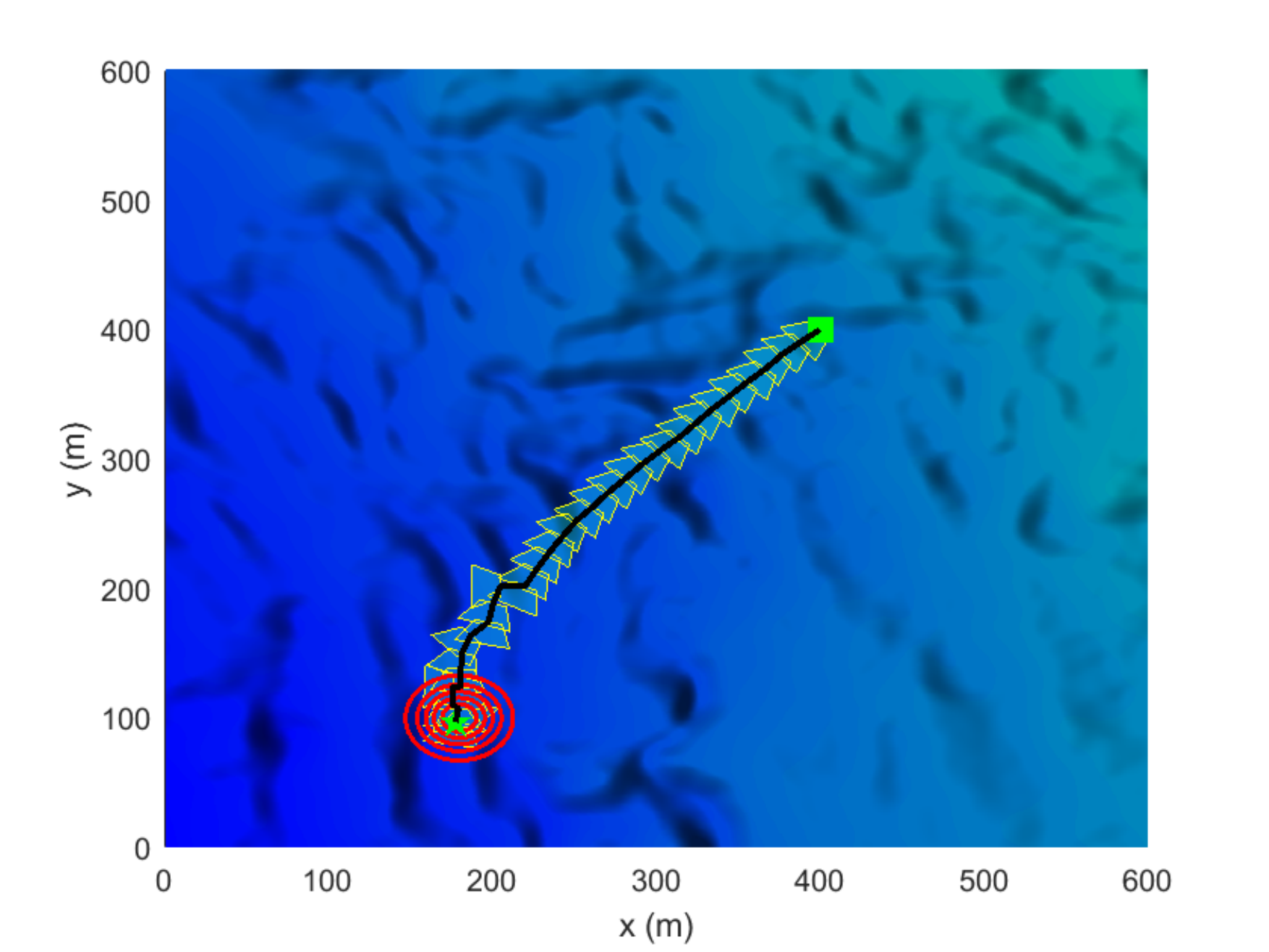}
    \caption{Scenario 1: J = 0.5291}
    \label{fig:scen1_2D}
    \end{subfigure}
    \begin{subfigure}[b]{0.45\textwidth}
    \centering
    \includegraphics[width=\textwidth]{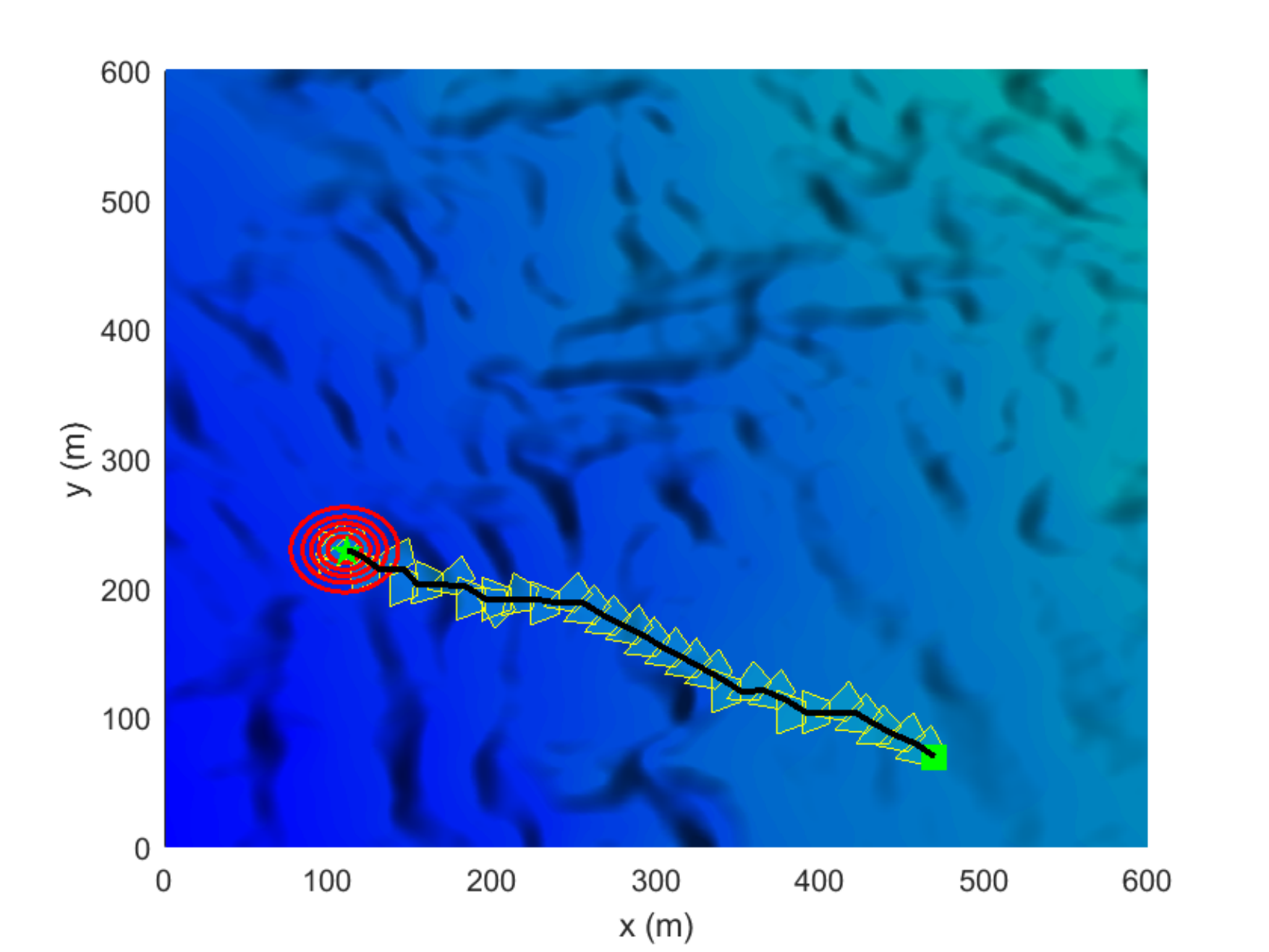}
    \caption{Scenario 2: J = 0.5009}
    \label{fig:scen2_2D}
    \end{subfigure}
    \begin{subfigure}[b]{0.45\textwidth}
    \centering
    \includegraphics[width=\textwidth]{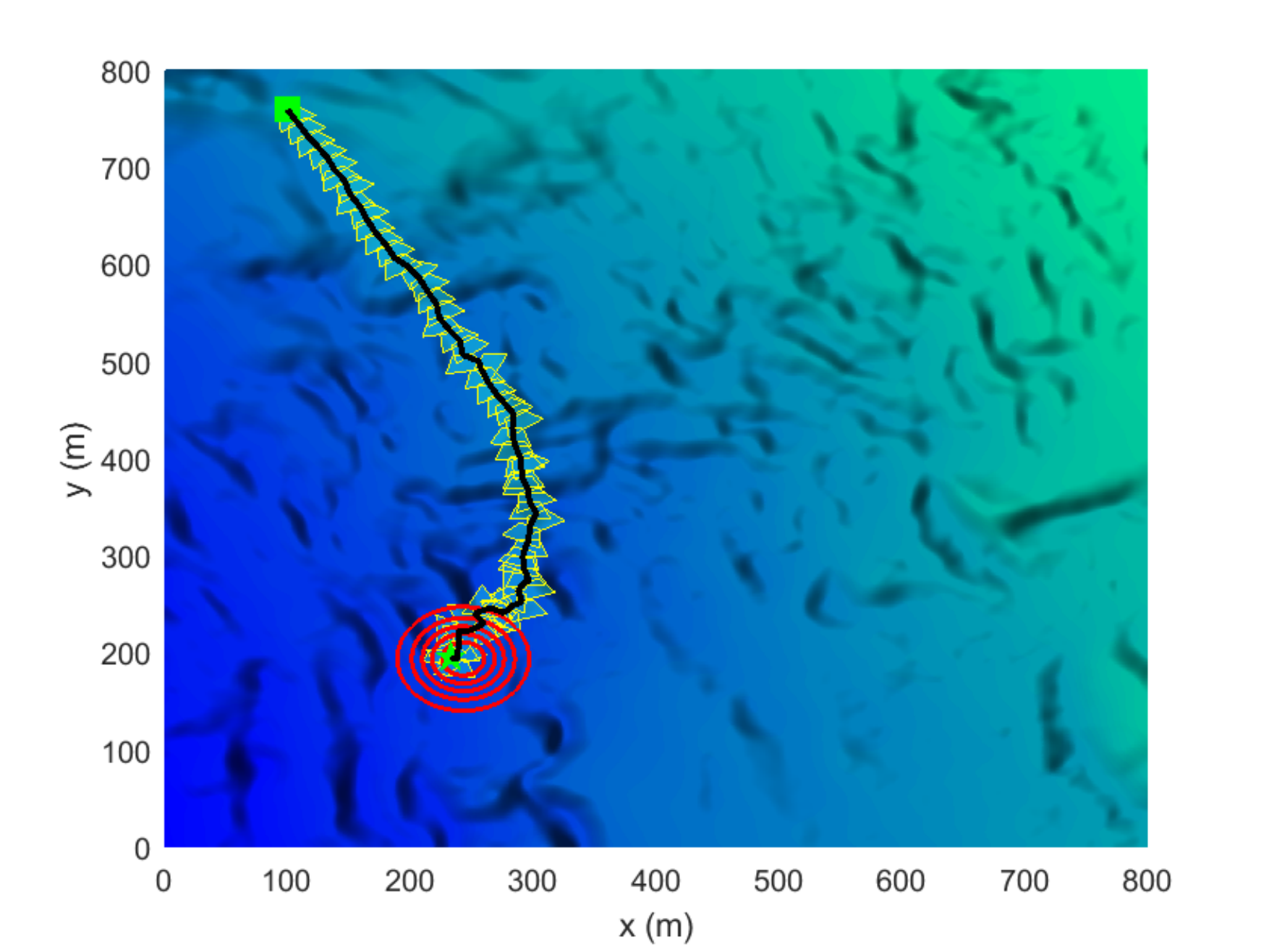}
    \caption{Scenario 3: J = 0.3086}
    \label{fig:scen3_2D}
    \end{subfigure}
    \begin{subfigure}[b]{0.45\textwidth}
    \centering
    \includegraphics[width=\textwidth]{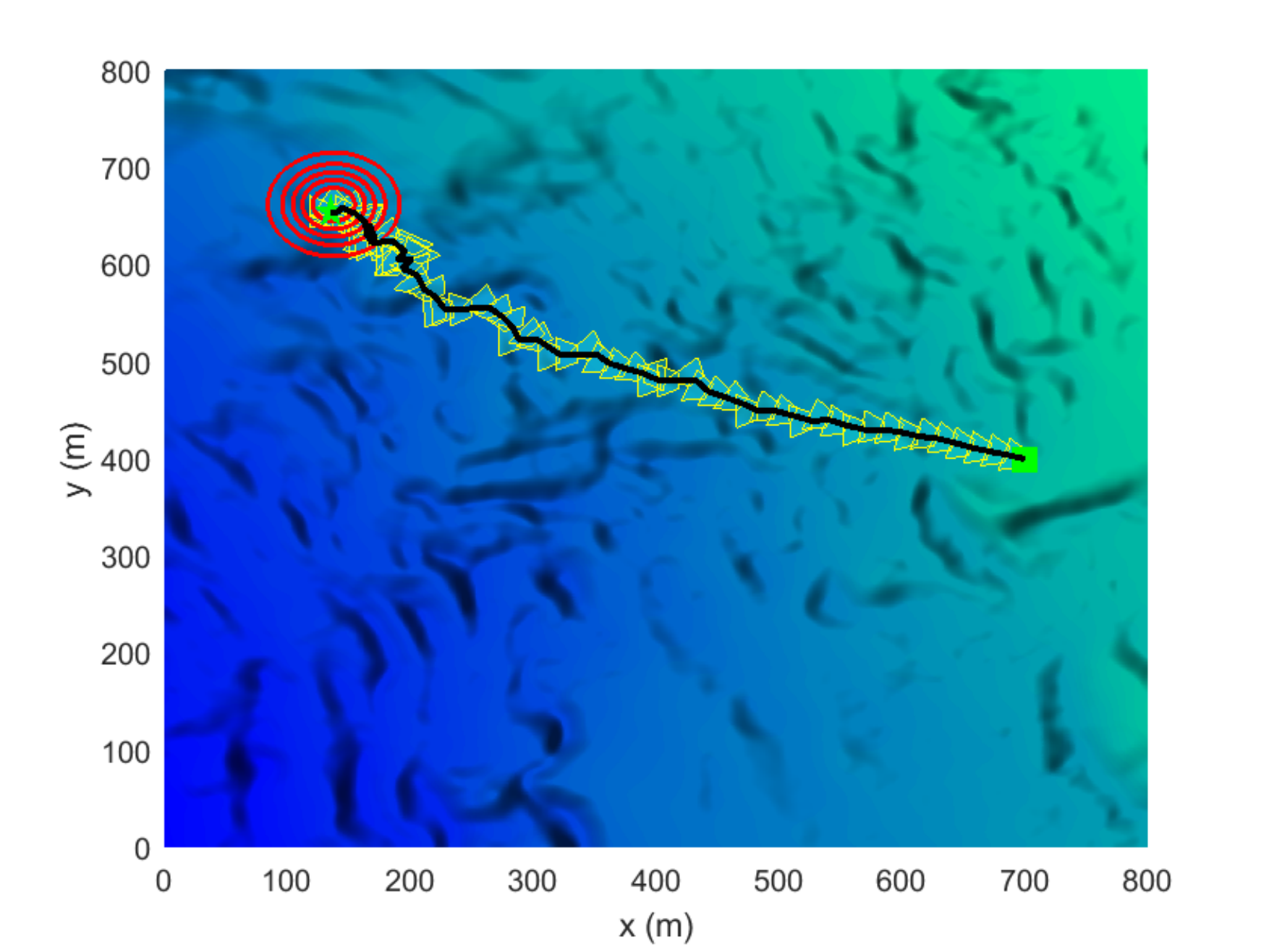}
    \caption{Scenario 4: J = 0.3209}
    \label{fig:scen4_2D}
    \end{subfigure}
    \begin{subfigure}[b]{0.45\textwidth}
    \centering
    \includegraphics[width=\textwidth]{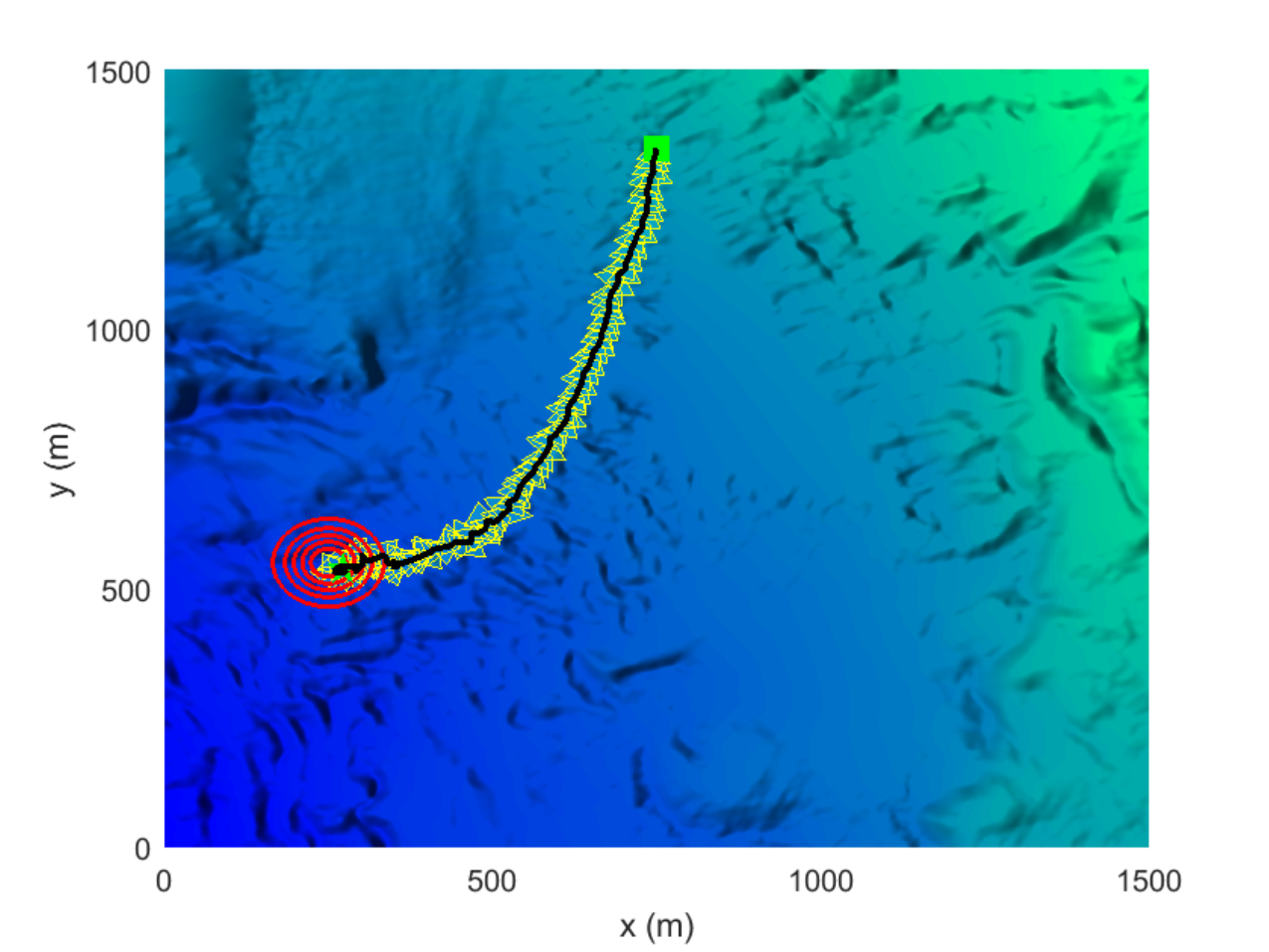}
    \caption{Scenario 5: J = 0.1257}
    \label{fig:scen5_2D}
    \end{subfigure}
    \begin{subfigure}[b]{0.45\textwidth}
    \centering
    \includegraphics[width=\textwidth]{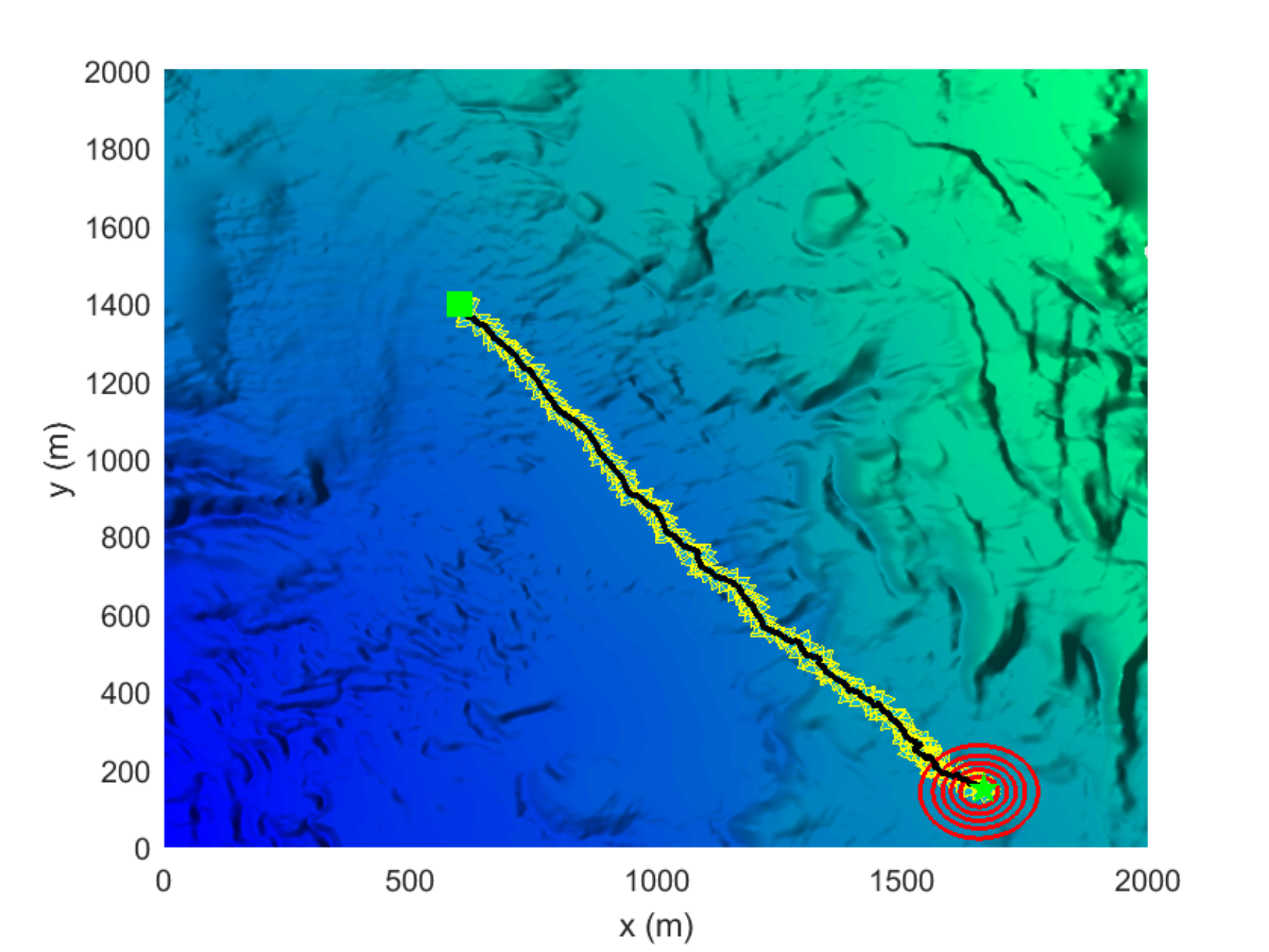}
    \caption{Scenario 6: J = 0.1064}
    \label{fig:scen6_2D}
    \end{subfigure}  
    \caption{Search paths generated by the proposed algorithm for each scenario}
    \label{fig:result_scen}
\end{figure*}

\begin{table*}
\centering
\caption{Detection probability under different levels of initial uncertainty}
\label{tbl:covariance}
\begin{tabular}{C{2.5cm} C{1.5cm} C{1.5cm} C{1.5cm} C{1.5cm} C{1.5cm} C{1.5cm}}
\hline \hline
Initial Uncertainty ($\sigma_0$) & Scenario 1 & Scenario 2 & Scenario 3 & Scenario 4 & Scenario 5 & Scenario 6\\ \hline \hline
10\% of map size & 0.9031	&0.8637  &0.8075	&0.7288	&0.5623  &0.4916 \\
25\% of map size & 0.7507	&0.7434  &0.6066	&0.6209	&0.4475	&0.3802 \\
50\% of map size & 0.5291	&0.5009	&0.3086	&0.3209	&0.1257	&0.1041 \\
\hline \hline
\end{tabular}
\end{table*}

The search paths generated for the UAV in 6 scenarios are depicted in Figure \ref{fig:result_scen}. The camera's footprints during the search are described as trapezoidal blocks with yellow borders. It can be seen that those flight paths guide the UAV to the area with the highest probability of the target's location. For instance, in Figure \ref{fig:scen6_2D}, the target moves towards the north of the map and the flight path gradually changes its direction at time $k = 60$ to track the target's movement. The algorithm is thus well adapted to the target's movement. Figure \ref{fig:result_scen} also shows that the algorithm is scalable in the sense that the paths always reach the areas with the highest probability of the target's location regardless of the size of the map and the number of waypoints. The algorithm therefore is sufficient for large-scale search operations.

In another evaluation, we examined the detection sensitivity with respect to uncertainty in the initial information. Three levels of uncertainty were considered with $\sigma_0$ set to 10\%, 25\%, and 50\% of the map size. The results in Table \mbox{\ref{tbl:covariance}} show a consistent trend across all scenarios: the detection probability decreases as $\sigma_0$ increases. This outcome is expected since greater uncertainty in the target's initial location spreads the belief over a larger area, thereby reducing the likelihood of detection. These results further support the validity of the proposed method.

\subsection{Comparison with other algorithms}

\begin{table}
\centering
\caption{Algorithm parameter settings}
\label{tbl:parameters}
\begin{tabular}{lll}
\toprule
\textbf{Algorithm} & \textbf{Parameter} & \textbf{Value} \\ \hline \hline 
\multirow{2}{*}{GA} & Crossover Percentage ($p_c$) & 0.4 \\
 & Mutation Percentage ($p_m$) & 0.15  \\ \hline 
\multirow{4}{*}{PSO} & Inertia weight ($w$) & 1 \\
& Inertia Weight Damping Ratio ($w_{damp}$) & 0.9 \\
 & Personal Learning Coefficient ($c_1$) & 1.0  \\
 & Global Learning Coefficient ($c_2$) & 1.0 \\ \hline 
GWO & Convergence parameter ($a$) & $2 \to 0$  \\ \hline 
\multirow{3}{*}{GSK} & Knowledge factor ($k_f$) & 0.5  \\
 & Knowledge ratio ($k_r$) & 0.5 \\
 & Dimension Junior ($D$) & 100  \\ \hline \hline 
\end{tabular}
\end{table}

\begin{figure*}
\centering
    \begin{subfigure}[b]{0.46\textwidth}
    \centering
    \includegraphics[width=\textwidth]{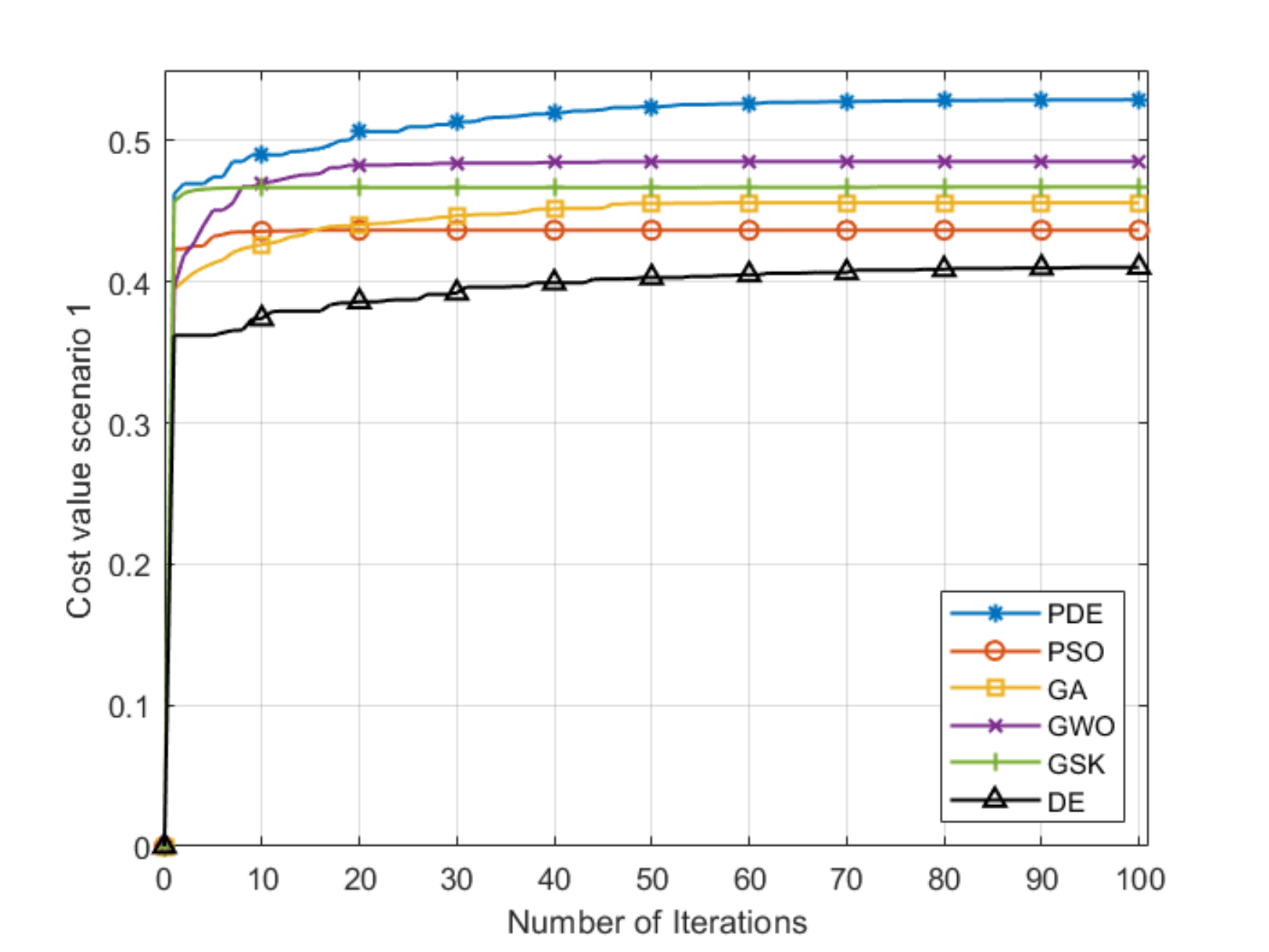}
    \caption{Scenario 1}
    \label{fig:scen1_cost}
    \end{subfigure}
    \begin{subfigure}[b]{0.46\textwidth}
    \centering
    \includegraphics[width=\textwidth]{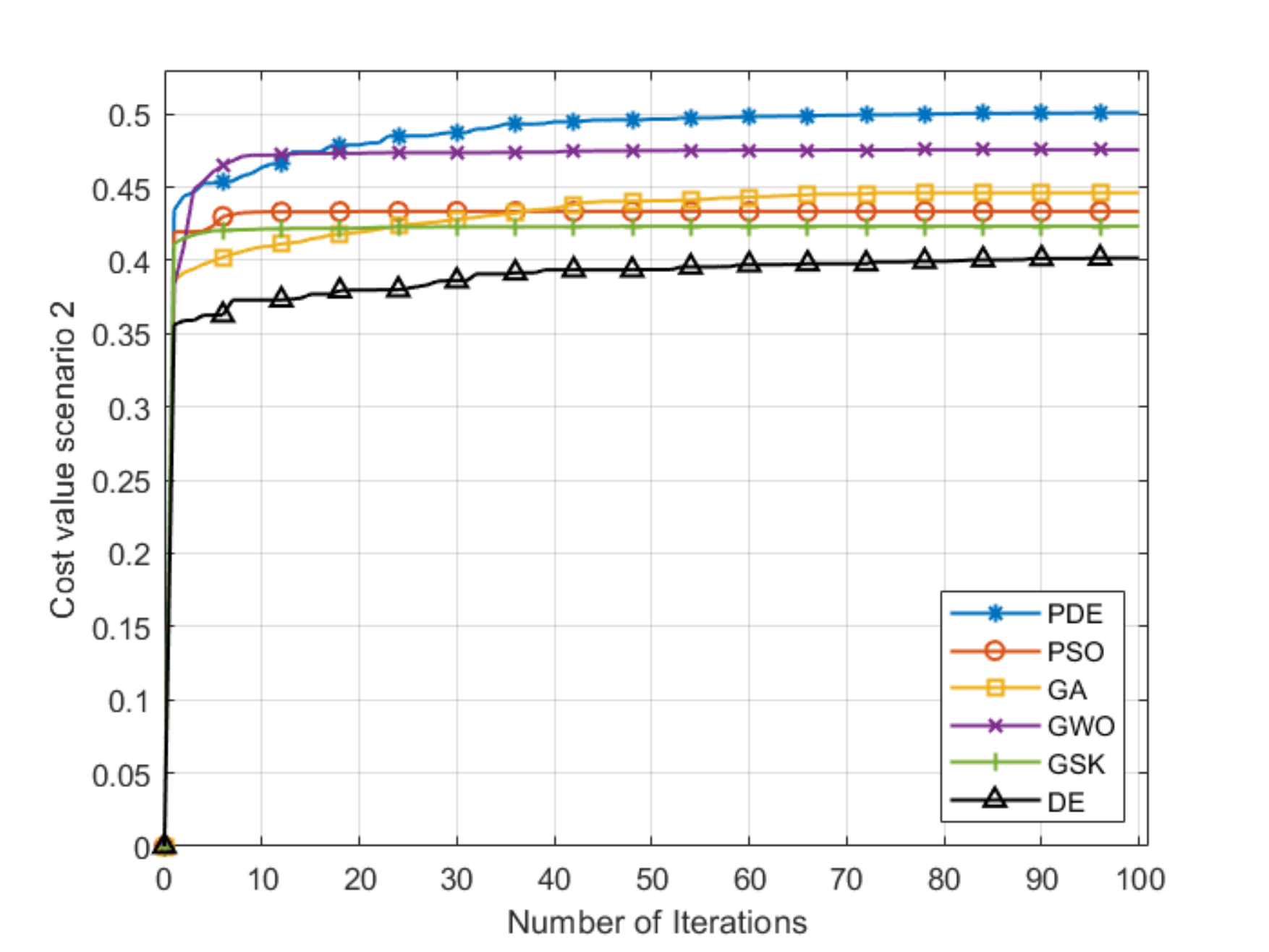}
    \caption{Scenario 2}
    \label{fig:scen2_cost}
    \end{subfigure}
    \begin{subfigure}[b]{0.46\textwidth}
    \centering
    \includegraphics[width=\textwidth]{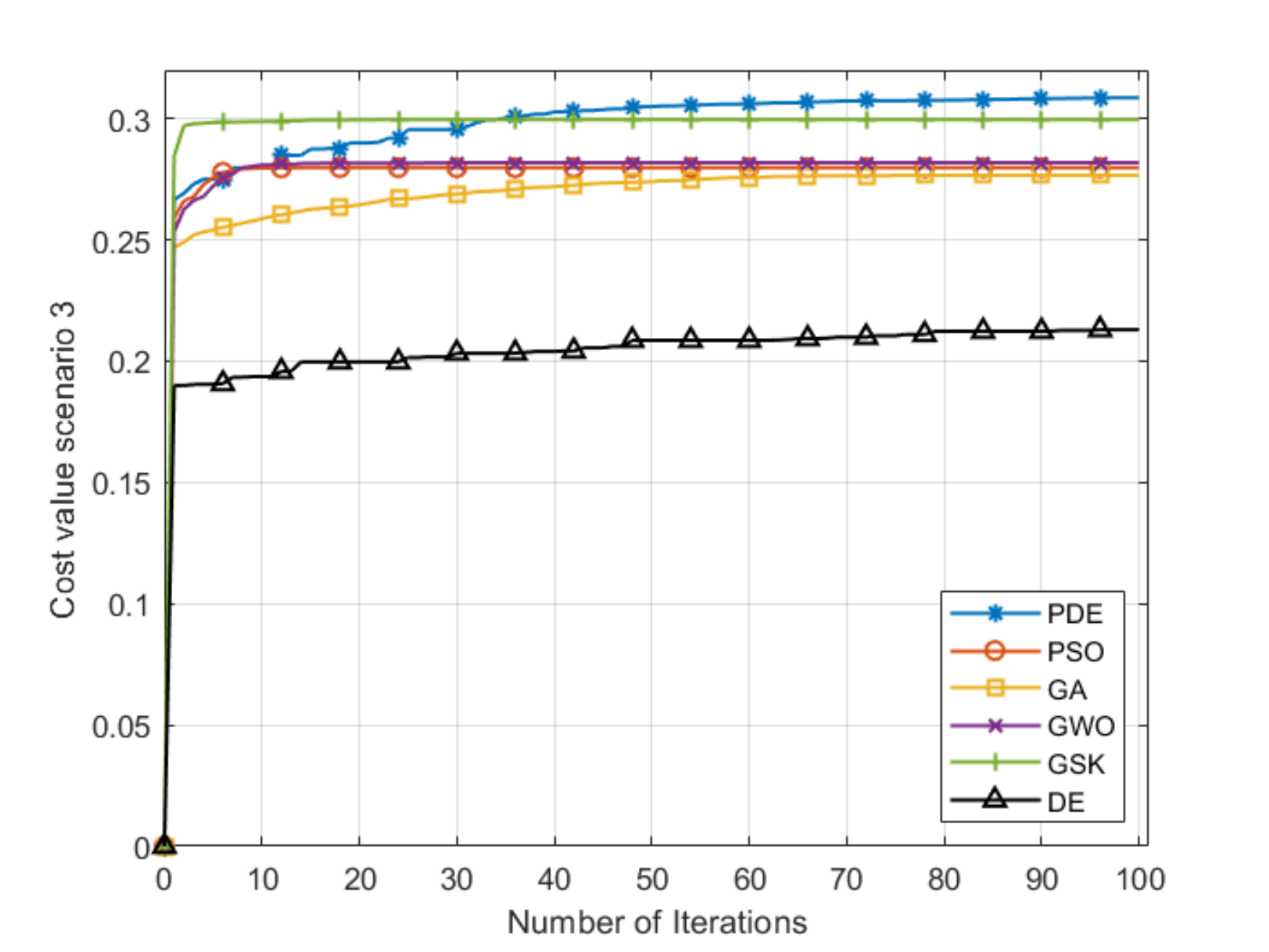}
    \caption{Scenario 3}
    \label{fig:scen3_cost}
    \end{subfigure}
    \begin{subfigure}[b]{0.46\textwidth}
    \centering
    \includegraphics[width=\textwidth]{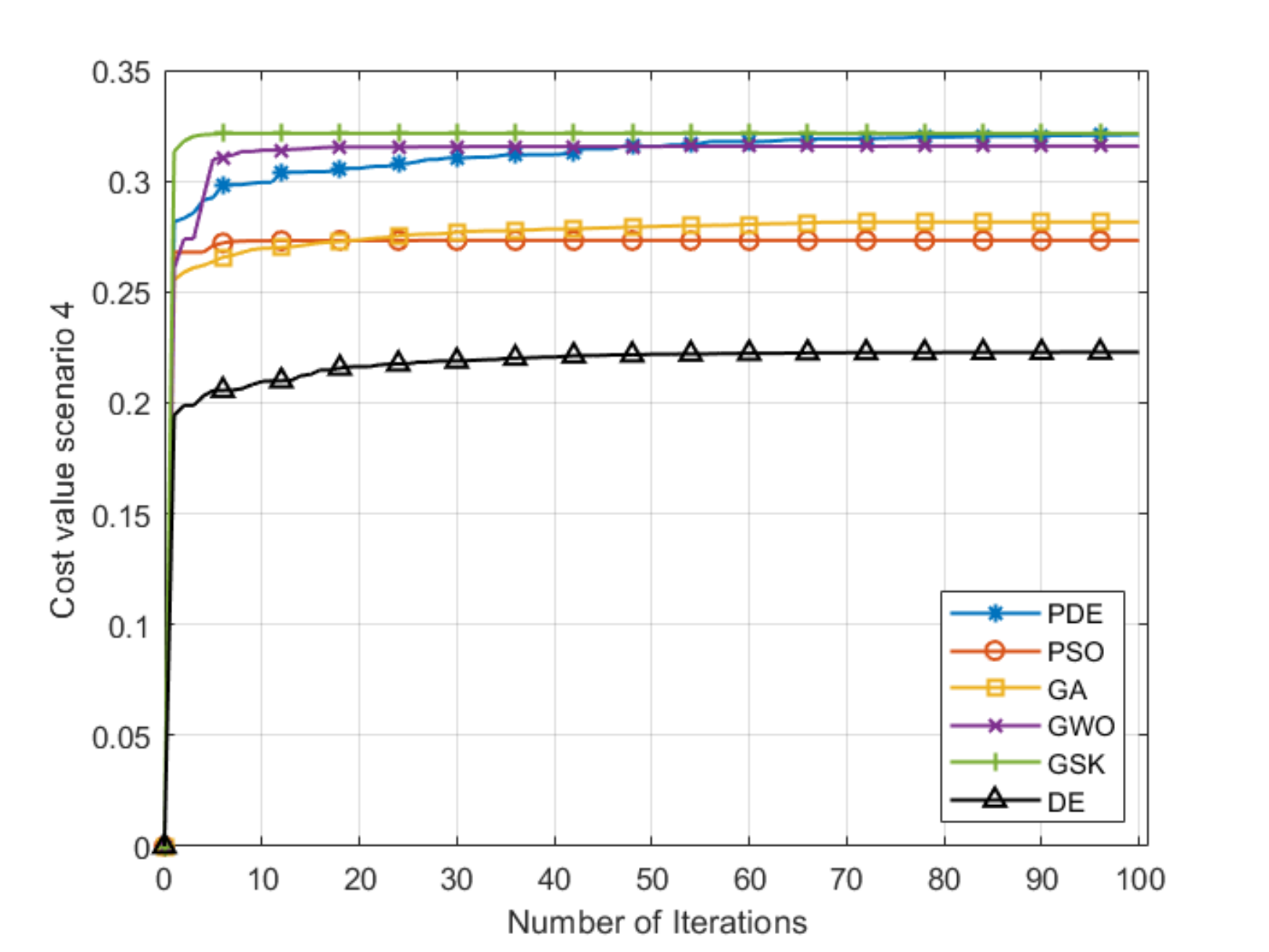}
    \caption{Scenario 4}
    \label{fig:scen4_cost}
    \end{subfigure}
    \begin{subfigure}[b]{0.46\textwidth}
    \centering
    \includegraphics[width=\textwidth]{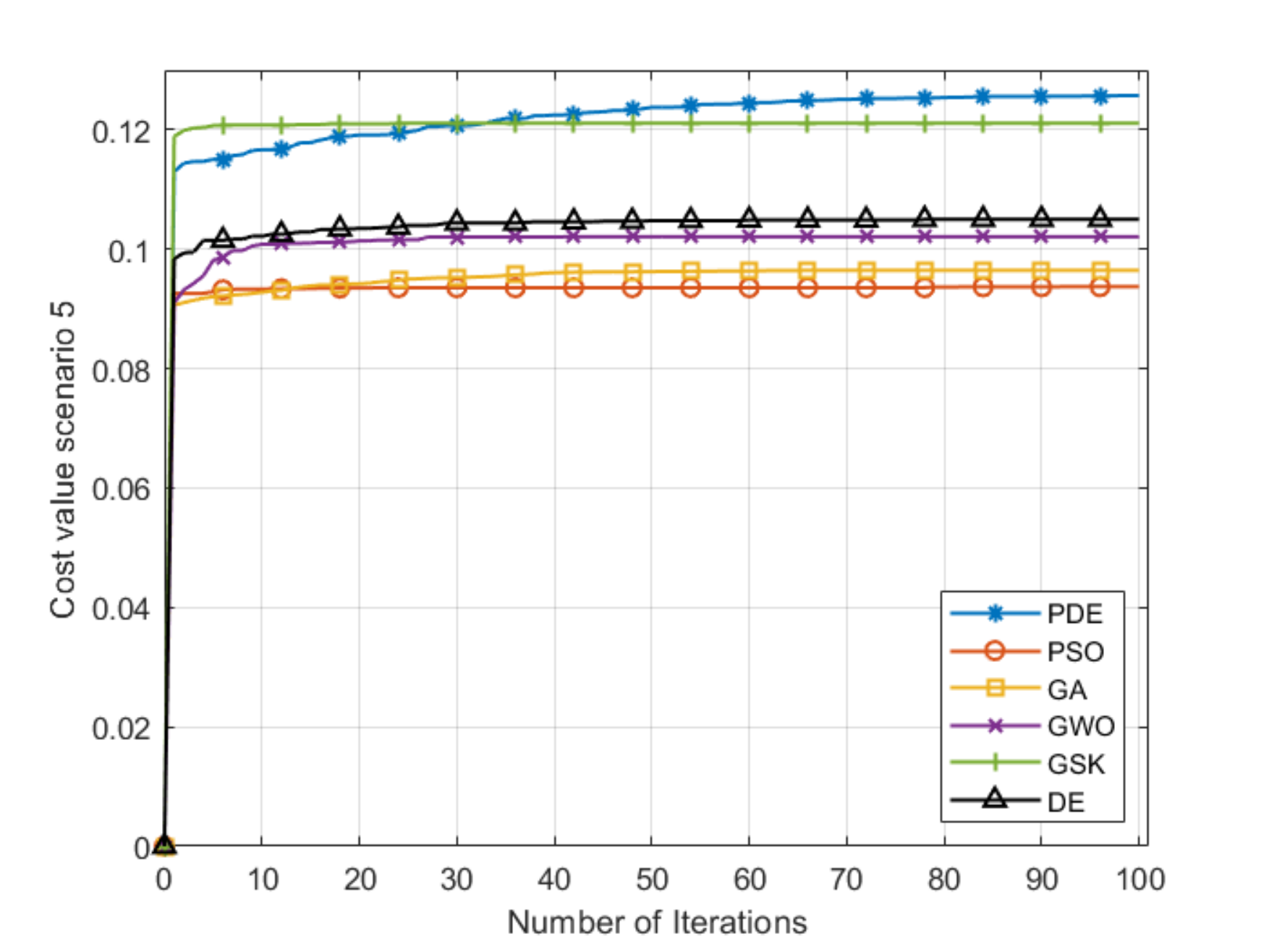}
    \caption{Scenario 5}
    \label{fig:scen5_cost}
    \end{subfigure}
    \begin{subfigure}[b]{0.46\textwidth}
    \centering
    \includegraphics[width=\textwidth]{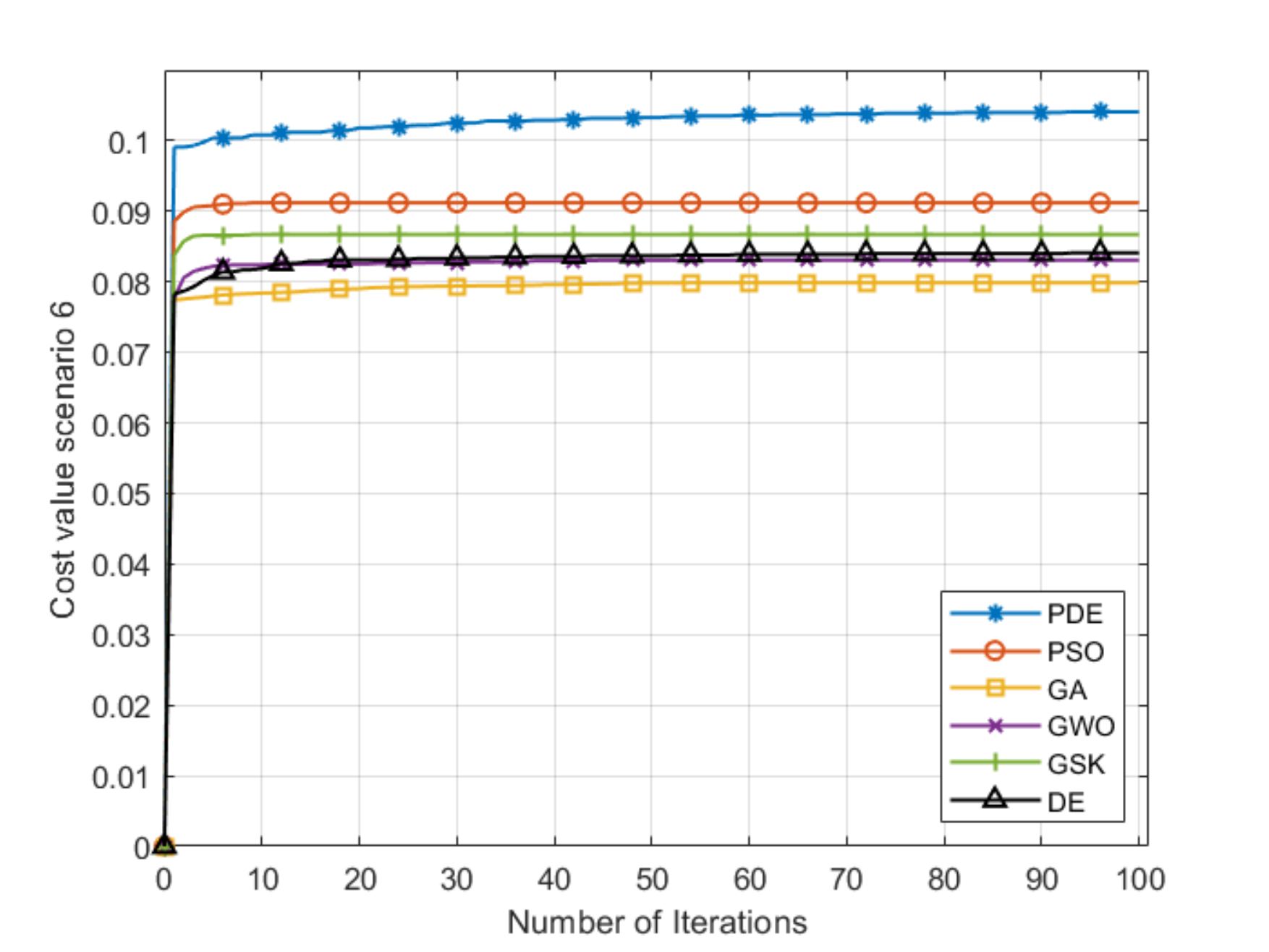}
    \caption{Scenario 6}
    \label{fig:scen6_cost}
    \end{subfigure} 
    \caption{Fitness values of the PDE and other metaheuristic algorithms on the six benchmark scenarios}
    \label{fig:cost_scen}
\end{figure*}

To further evaluate the performance of the proposed algorithm, we conducted comparisons with other state-of-the-art metaheuristic algorithms including the standard differential evolution (DE)~\cite{Storn1997, pant2020differential}, genetic algorithm (GA)~\cite{538609,Katoch2020}, particle swarm optimization (PSO)~\cite{488968,Gad2022}, grey wolf optimization (GWO)~\cite{Mirjalili2014, NadimiShahraki2021}, and gaining-sharing knowledge algorithm (GSK)~\cite{Mohamed2019}. The evaluation metrics used for comparison include: (i) the fitness value, which represents the cumulative probability of finding the target as defined in \eqref{eqn:cost}, and (ii) the execution time, which reflects the algorithm's efficiency. Table \mbox{\ref{tbl:parameters}} lists the parameter settings used for each algorithm. For a fair comparison, all metaheuristic methods were implemented using the same population size and maximum number of iterations as those used for the proposed PDE. Due to the stochastic nature of those methods, each method is executed 10 times, and the average values are used for comparison.

\begin{table*}
\centering
\caption{Comparison on the fitness (Eq. \ref{eqn:cost})}
\label{tbl:compare_cost}
\begin{tabular}{C{0.5cm} C{2.35cm} C{2.35cm} C{2.35cm} C{2.35cm} C{2.35cm} C{2.35cm}}
\hline \hline
Scen. & PDE & PSO & GA & GWO & GSK & DE\\ \hline \hline
1 & $\mathbf{0.5291\pm0.001}$ & $0.4365 \pm 0.006$ & $0.4561 \pm 0.015$ &$0.4852 \pm 0.014$ & $0.4672 \pm 0.014$ & $0.4103 \pm 0.010$  \\
2 & $\mathbf{0.5009 \pm 0.014}$ &$0.4335 \pm 0.002$ & $0.4461 \pm 0.009$ & $0.4756 \pm 0.012$ &$0.4233 \pm 0.036$ & $0.4016 \pm 0.007$ \\
3 & $\mathbf{0.3086 \pm 0.003}$ & $0.2797 \pm 0.005 $ & $0.2766 \pm  0.003$& $0.2818 \pm  0.004$ &$0.2996 \pm 0.023$ &$0.2130 \pm 0.016$ \\
4 & $0.3209  \pm 0.001$ & $0.2732 \pm 0.001$ & $0.2815 \pm 0.002$ & $0.3157 \pm 0.007$& $\mathbf{0.3215 \pm 0.011}$ & $0.2228 \pm 0.005$  \\
5& $\mathbf{0.1257 \pm 0.007}$ & $0.0937 \pm 0.002$ & $0.0964 \pm 0.003$ & $0.1020 \pm 0.006$ & $0.1211 \pm 0.015$ &$0.1050 \pm 0.003$  \\
6& $\mathbf{0.1041 \pm 0.002}$ & $0.0912 \pm 0.001$ &$0.0798 \pm 0.007$ &$0.0830 \pm 0.007$ & $0.0867 \pm 003$ &$0.0841 \pm 0.008$\\ \hline \hline
\end{tabular}
\end{table*}

\begin{table*}
\centering
\caption{Comparison on the execution time (unit: seconds)}
\label{tbl:compare_time}
\begin{tabular}{C{1cm} C{2cm} C{2cm} C{2cm} C{2cm} C{2cm} C{2cm}}
\hline \hline
Scen. & PDE & PSO & GA & GWO & GSK & DE\\ \hline \hline
1 & $\mathbf{246 \pm 16}$ & $269 \pm 21$ & $249 \pm 33$ &$276\pm 47$ & $564 \pm 74$& $299 \pm 20$ \\
2 & $\mathbf{234 \pm 28}$ & $252 \pm 19$ & $241 \pm 9$& $237 \pm 24$& $714 \pm 72$ & $307 \pm 11$  \\
3 & $516 \pm 8$ &$494 \pm 23$ &$505 \pm 11$ & $\mathbf{469} \pm 24$& $1467 \pm 36$& $501 \pm 34$ \\
4 & $\mathbf{523 \pm 43}$ & $736 \pm 65$ & $561 \pm 34$ & $576\pm 62$ & $1791 \pm 59$ & $607 \pm 28$  \\
5 & $\mathbf{907 \pm 85}$ &$1066 \pm 95$ &$956 \pm 57$ & $1007 \pm 112$ &$2841 \pm 43$ & $1005 \pm 58$ \\ 
6 & $\mathbf{1668 \pm 83}$ & $1760 \pm$ 70& $1677 \pm 71$& $1702 \pm 39$ & $2944 \pm 29$ & $1754 \pm 63$ \\
\hline \hline
\end{tabular}
\end{table*}

Table \ref{tbl:compare_cost} shows the fitness values obtained by the algorithms. It can be seen that the PDE outperforms other algorithms in detection probability with the best results in 5 out of 6 scenarios. In Scenario 4, where the GSK achieves the best result, its performance is only 0.2\% better than that of the PDE, and thus does not indicate a decline in the PDE's performance because the PDE still outperforms all other algorithms. Instead, this suggests that the knowledge-sharing mechanism in GSK is particularly effective at this specific scale. The original DE, on the other hand, performs poorly in most scenarios due to its use of the Cartesian coordinate. Especially, it is trapped at local optima in scenarios 3 and 4 with low fitness values as shown in figures \ref{fig:scen3_cost} and \ref{fig:scen4_cost}. The results thus show that the use of the polar coordinate significantly improves the search performance of the PDE algorithm. 

Table \mbox{\ref{tbl:compare_cost}} also shows a clear trend that the detection probability decreases as the map size increases. This result is expected because enlarging the search area spreads the target belief over a larger region, thereby reducing the probability of detection within any given area. These results highlight the need for multiple coordinated UAVs to enable scalable target detection in large environments.

In another evaluation, Table \ref{tbl:compare_time} shows the execution time of the algorithms on the same hardware with an Intel Core i7-8565U, 1.8GHz processor. It can be seen that the PDE is fastest in 5 out of 6 scenarios. To understand this efficiency gain, it is necessary to examine the computational complexity of the search process. In all tested algorithms, the dominant computational cost is the fitness evaluation $J$, which requires updating the Bayesian belief map over a discrete search grid. For a search map discretized into $M \times M$ cells and a path containing $N$ waypoints, evaluating the fitness of a single candidate path carries a complexity of $\mathcal{O}(N \cdot M^2)$. Given a population size of $N_p = 500$ and $100$ iterations, the fitness function is called $50,000$ times per run. Because both the standard Cartesian DE and the proposed PDE perform this dominant operation an identical number of times, the observed difference in execution time originates from the per-call candidate generation and constraint handling overhead.

In the Cartesian DE, formulating the camera footprint requires extracting the UAV's heading angle $\psi$ and step length $\rho$ from raw $(x,y)$ coordinates. This process requires computationally expensive trigonometric functions (e.g., $\text{atan2}$) and square root operations at every node. Furthermore, handling physical kinematic bounds such as maximum velocity or turning limits in Cartesian space requires complex constraint calculations. Across $50,000$ evaluations of an $N$-node path (e.g., $N=200$ in Scenario 6), this introduces tens of millions of redundant trigonometric operations. In contrast, the proposed PDE natively represents candidate paths using $\rho$ and $\psi$. Kinematic constraints are enforced via simple scalar clamping, $\rho \in [\rho_\text{min}, \rho_\text{max}]$, and the heading angle is passed directly to the footprint evaluation without conversion. while the heading angle is used directly in the footprint evaluation without requiring additional coordinate conversion. This design helps reduce computational overhead and explains the faster execution time of the PDE compared with Cartesian-based algorithms.

\subsection{Validation on a real UAV platform}
To evaluate the validity of the proposed PDE algorithm for real UAVs, we have carried out experiments with details as follows.

\subsubsection{Experimental setup}
\begin{figure*}
    \centering
    \begin{subfigure}[b]{0.47\textwidth}
    \centering
    \includegraphics[width=\textwidth]{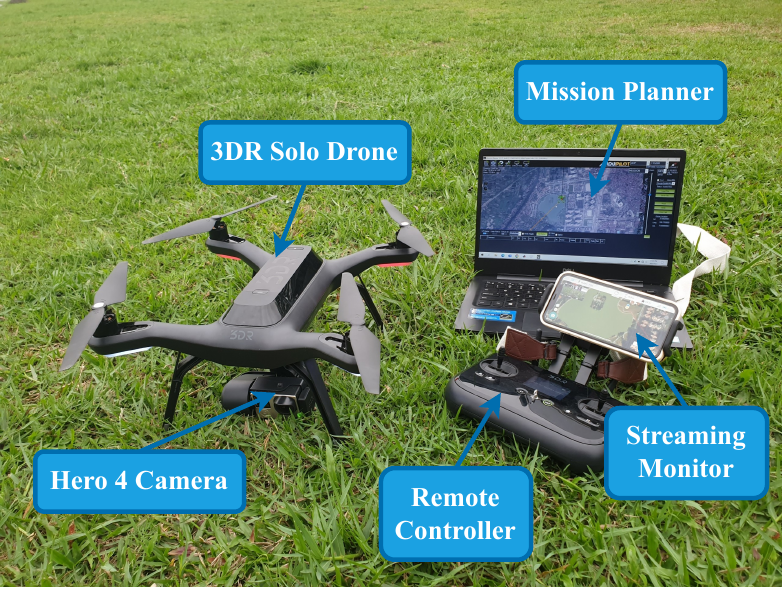}
    \caption{The 3DR Solo Drone and its flight controller}
    \label{fig:devices}
    \end{subfigure}
    \begin{subfigure}[b]{0.51\textwidth}
    \centering
    \includegraphics[width=\textwidth]{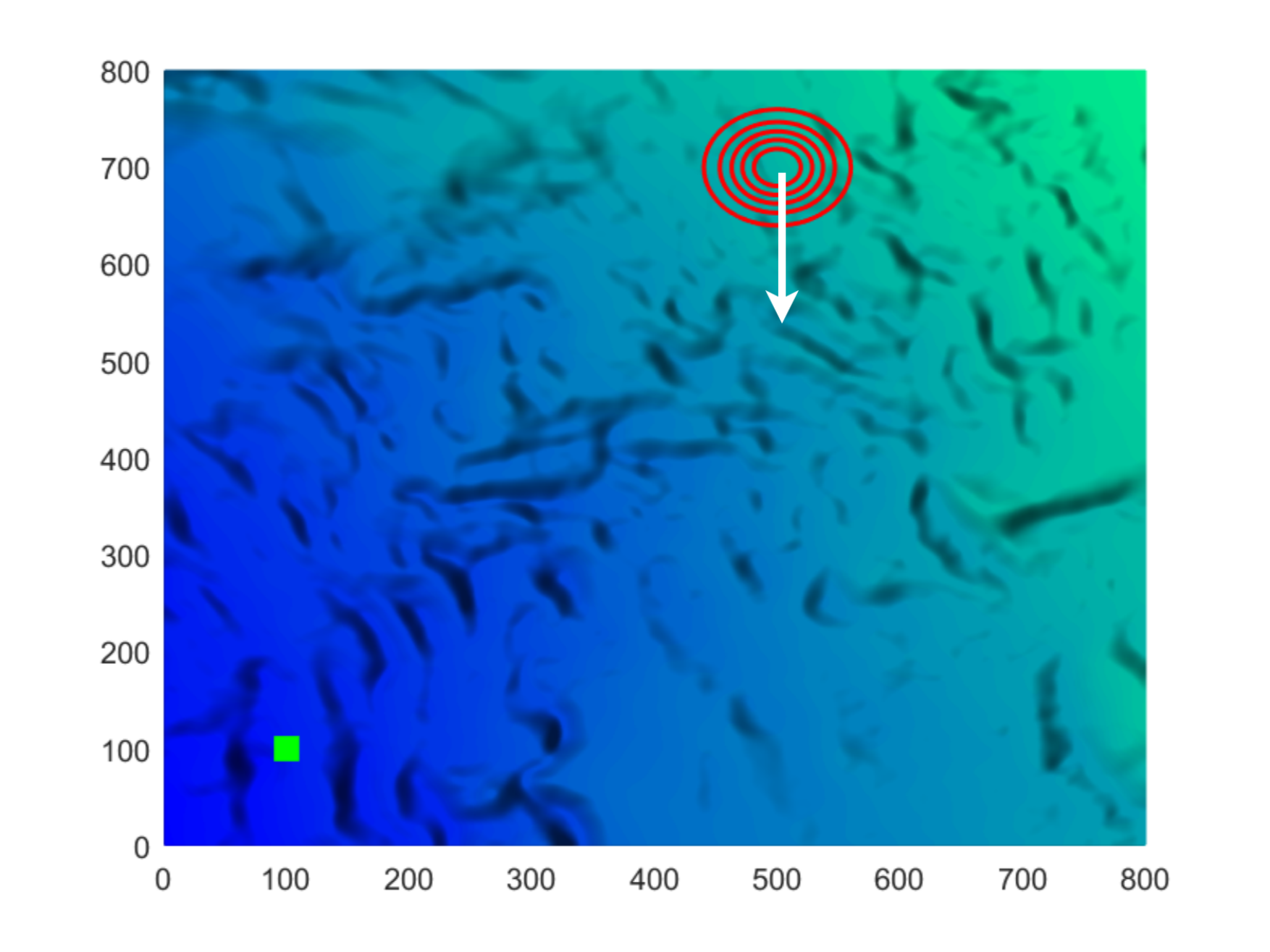}
    \caption{The initial belief map}
    \label{fig:sim_init}
    \end{subfigure}
    \caption{Experimental setup}
\end{figure*}

The experiment is conducted within an area of 80 m $\times$ 80 m. A 3DR Solo drone, which can be controlled via a ground control station (GCS) software named Mission Planner, is used as depicted in Figure \ref{fig:devices}. The drone is equipped with a Hero 4 camera functioning as a visual sensor. Its horizontal and vertical angles of view are $\alpha=80^\circ$ and $\beta=60^\circ$, respectively. A three-axis gimbal is used to stabilize and control the camera to look downward at a fixed angle of $\phi=30^\circ$. When the UAV operates, the video captured by the camera is streamed to the GCS for target detection. The detection is carried out by an observer having the target task performance $\text{TTP} = 47.327$~\cite{vollmerhausen2004new} and the critical period $V_{50}=2.7$~\cite{espinola2007modeling}. The target is a pedestrian moving in the southern direction at a speed of around 1.5 m/s. Her last known location is at a latitude of 21.0528321 and a longitude of 105.7764591, which is then used as the input to create an initial belief map. In this map's coordinate, the initial position of the UAV is $(10,10)$ (green square) and the last known location of the target is set to $(50,70)$, as shown in Figure \ref{fig:sim_init}.
\begin{table}
\centering
\caption{Parameters used in experiments}
\label{tbl:ex_param}
\begin{tabular}{p{4cm} C{1.5cm} C{1.5cm}}
\hline \hline
\multicolumn{1}{c}{Parameter} & Notation & Value\\ \hline
Horizontal angles & $\alpha$ & $80^\circ$ \\
Vertical angles & $\beta$ &  $60^\circ$\\ 
Gimbal angle & $\phi$ & $30^\circ$ \\ 
Flight height & $h$ & 6.5 m \\ 
Target tracking performance & TTP & 47.327 \\ 
Critical period & $V_{50}$  & 2.7 \\ 
Area of target & $S^t$ & 0.25 m$^2$ \\ 
Speed of target & & 1.5 m/s \\ 
UAV maximum speed & $\rho_\text{max}$ & 5 m/s \\ 
 \hline \hline
\end{tabular}
\end{table}

During experiments, the maximum speed of the drone is set to 5 m/s and the flight altitude is 6.5~m above the ground. At this altitude, the footprint of the camera covers an area of 30.42  m$^2$, which is suitable for the relatively small search area designated for this experiment. Detailed parameters for experiments are summarized in Table {\ref{tbl:ex_param}}. 

\begin{figure*}
\centering
    \begin{subfigure}[b]{0.525\textwidth}
    \centering
    \includegraphics[width=\textwidth]{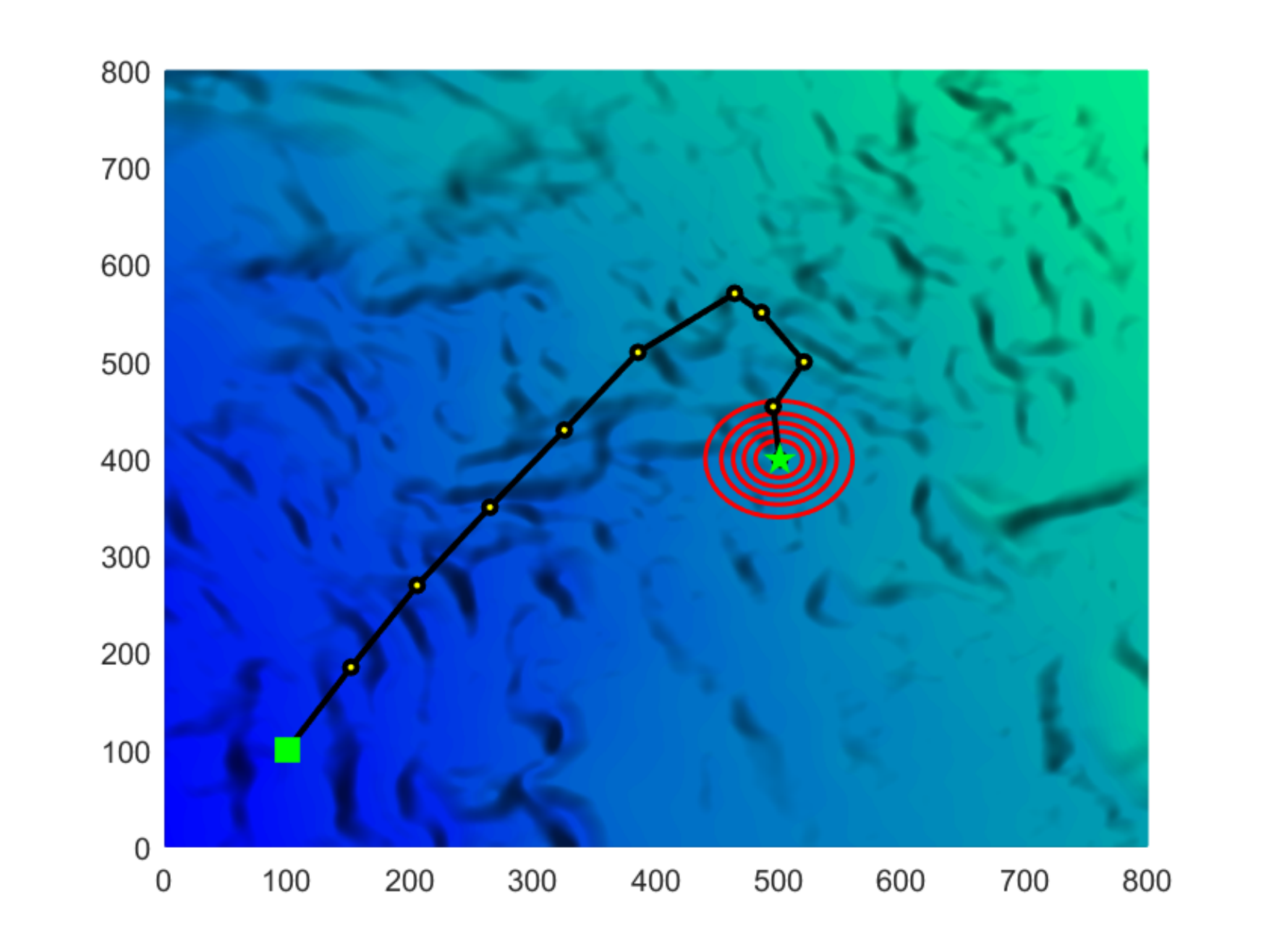}
    \caption{The planned search path in the belief map}
    \label{fig:sim_path}
    \end{subfigure}
    \begin{subfigure}[b]{0.45\textwidth}
    \centering
    \includegraphics[width=\textwidth]{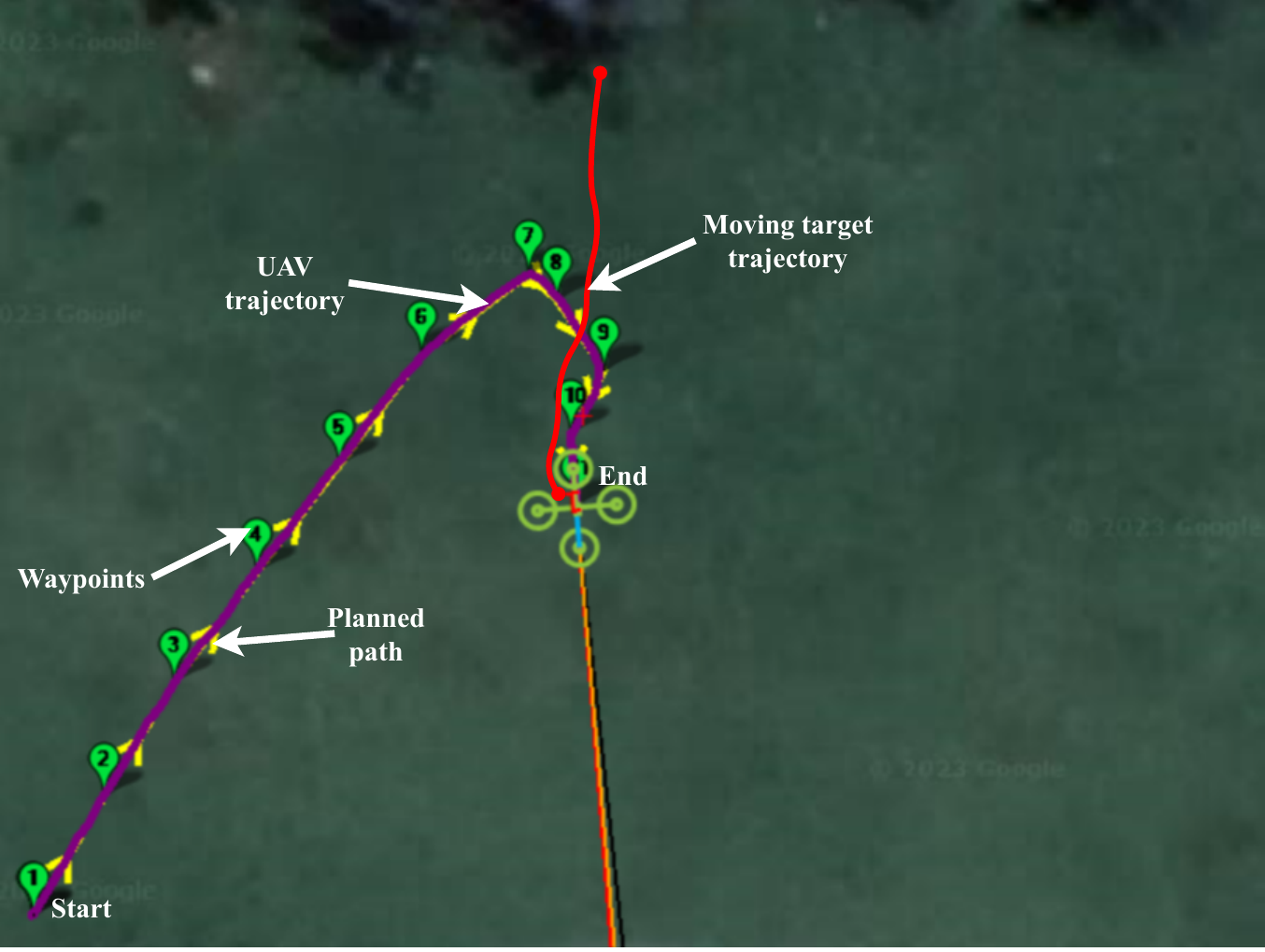}
    \caption{The planned (yellow) and actual (violet) flight paths together with the target's moving path (red)}
    \label{fig:exp_detect}
    \end{subfigure}
    \caption{Experimental results}
    \label{fig:tracking}
\end{figure*}

\begin{figure*}
    \centering
    \includegraphics[width=0.55\textwidth]{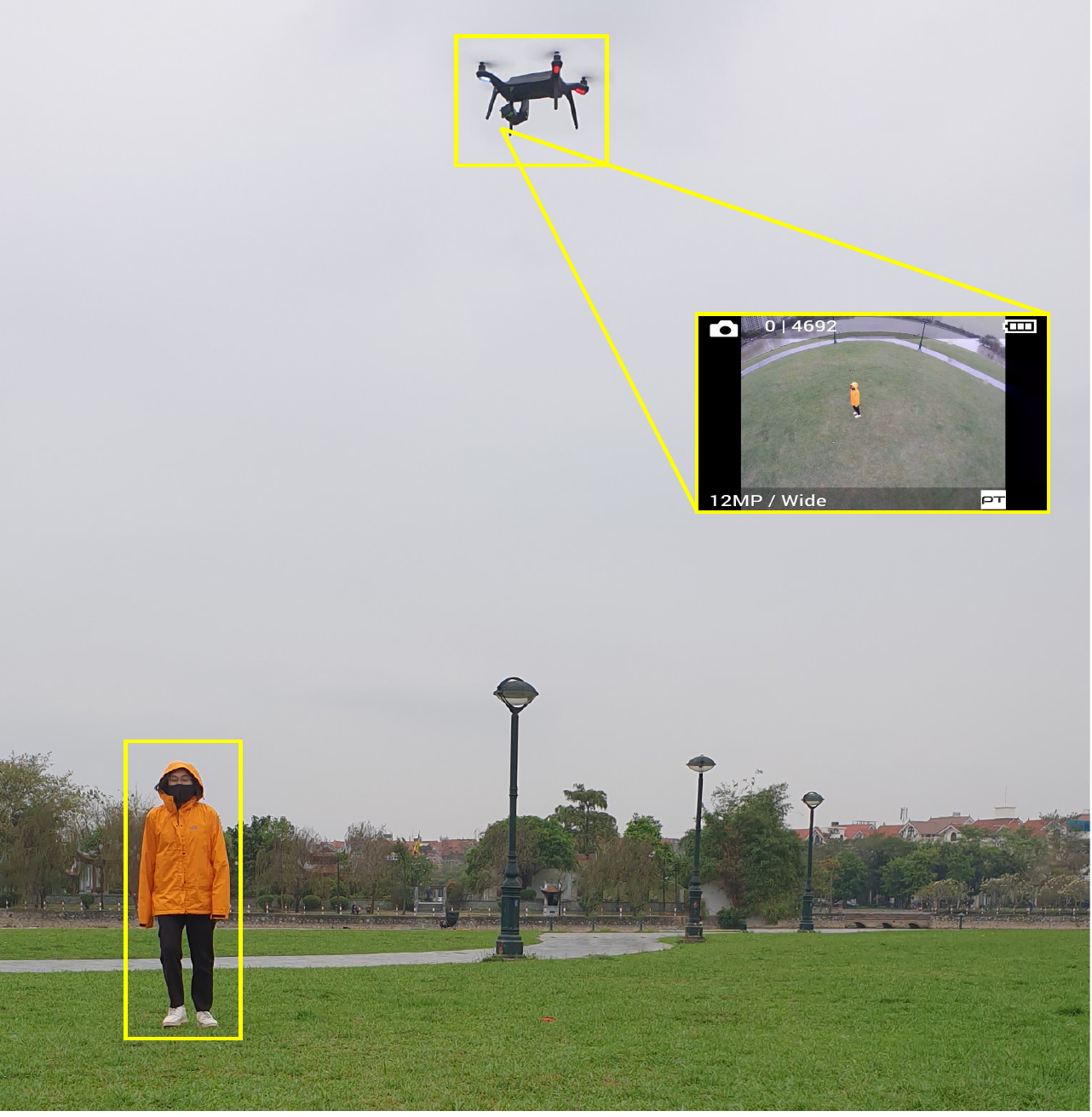}
    \caption{The target within the camera's field of view}
    \label{fig:experiment}
\end{figure*}

\subsubsection{Experimental results}
Figure \mbox{\ref{fig:sim_path}} shows the search path consisting a sequence of waypoints generated by the PDE algorithm implemented in MATLAB. This path is subsequently converted into geographic coordinates and uploaded to the UAV using Mission Planner software. The UAV flight controller then smooths the path to satisfy kinematic constraints, such as the mininum turning radius, before passing it to the low-level tracking controller. Figure \ref{fig:exp_detect} shows the planned (yellow) and the actual (violet) flight paths recorded through Mission Planner. The path of the target is depicted by the red line. The overlap between the actual and the planned paths means that the path generated by the PDE is feasible for the UAV to follow. It also can be seen that the UAV approaches the target at (21.0527534, 105.7764597) when it is within the camera's field of view. Figure \ref{fig:experiment} shows the target and its appearance in the photo captured by the UAV. Note that the detection event happened in this experiment is due to the alignment of our setup with the assumptions used in theory, e.g., the target actually started at the last known location and then moved along the predetermined direction. In reality, this may not be the case and hence there is no guarantee that the UAV will find the target. Nevertheless, the experimental results confirmed the feasibility of the generated path for the UAV to follow and the validity of the proposed approach in maximizing the probability of finding the target.

\section{Conclusion}
In this work, we have presented a polar coordinate-based differential evolution (PDE) algorithm for the search problem with a focus on increasing the probability of finding a dynamic target and reducing computation time. The proposed algorithm estimates the target's location based on a geometric analysis of the camera's field of view and the target's observation probability. It incorporates constraints associated with the UAV kinematics and target dynamics to narrow the search space and generate feasible solutions. The use of polar coordinates enhances the algorithm's robustness to local optima and leads to improved solution quality. The Gaussian belief map combined with Bayesian updates mitigates motion uncertainty and measurement noise, while the NVESD metric quantifies target visibility under non-ideal sensing conditions. The effectiveness of the PDE algorithm is evaluated through simulations and comparisons in various search scenarios. Experiments have also been conducted to confirm the practical use of the algorithm. Due to physical limitations, such as battery capacity and allowable path length, a single UAV can only cover a limited area within a given time horizon, which in turn limits the probability of detection. To address this limitation, future work will focus on extending the proposed framework to coordinated multi-UAV systems for target search in large-scale dynamic environments.

\section*{Acknowledgment}
Thu Hang Khuat was funded by the Master, PhD Scholarship Programme of Vingroup Innovation Foundation (VINIF), code VINIF.2024.Ths.16.

\bibliographystyle{ieeetr}  
\bibliography{mybibfile2}
\balance

\end{document}